\documentclass{article} 
\usepackage{iclr2023_conference,times}
\usepackage{preamble}

\usepackage{hyperref}
\usepackage{url}

\title{%
Weighted Ensemble Self-Supervised Learning
}

\author{%
{\bf Yangjun Ruan\thanks{University of Toronto \& Vector Institute. Work done as a student researcher at Google.}~~\thanks{Correspondence to \texttt{yjruan@cs.toronto.edu}, \texttt{\{iansf,jvdillon\}@google.com}.}\quad\quad Saurabh Singh\quad\quad Warren Morningstar\quad\quad Alexander A.~Alemi \\
Sergey Ioffe\quad\quad Ian Fischer\footnotemark[2]\quad\quad Joshua V. Dillon\footnotemark[2]}\\
Google Research\par
}

\iclrfinalcopy 
\begin{document}

\maketitle

\vspace{-1.5\baselineskip}
\begin{abstract}
\vspace{-.5\baselineskip}
Ensembling has proven to be a powerful technique for boosting model performance, uncertainty estimation, and robustness in supervised learning.
Advances in self-supervised learning (SSL) enable leveraging large unlabeled corpora for state-of-the-art few-shot and supervised learning performance.
In this paper, we explore how ensemble methods can improve recent SSL techniques by developing a framework that permits data-dependent \emph{weighted} cross-entropy losses.
We refrain from ensembling the representation backbone; this choice yields an efficient ensemble method that incurs a small training cost and  requires no architectural changes or computational overhead to downstream evaluation.
The effectiveness of our method is demonstrated with two state-of-the-art SSL methods, DINO \citep{caron2021dino} and MSN \citep{assran2022msn}.
Our method outperforms both in multiple evaluation metrics on ImageNet-1K, particularly in the few-shot setting.
We explore several weighting schemes and find that those which increase the diversity of ensemble heads lead to better downstream evaluation results.
Thorough experiments yield improved prior art baselines which our method still surpasses; \eg, our overall improvement with MSN \vitb/16 is 3.9 \pp for 1-shot learning.
\end{abstract}
\vspace{-\baselineskip}
\section{Introduction}
\vspace{-.5\baselineskip}
\label{sec:intro}

\definecolor{steelblue}{rgb}{0.27, 0.51, 0.71}
\definecolor{babyblueeyes}{rgb}{0.63, 0.79, 0.95}
\definecolor{carolinablue}{rgb}{0.6, 0.73, 0.89}
\definecolor{columbiablue}{rgb}{0.61, 0.87, 1.0}
\definecolor{lightskyblue}{rgb}{0.53, 0.81, 0.98}
\definecolor{sinopia}{rgb}{0.8, 0.25, 0.04}
\definecolor{cornellred}{rgb}{0.7, 0.11, 0.11}
\definecolor{darkpastelred}{rgb}{0.76, 0.23, 0.13}
\definecolor{debianred}{rgb}{0.84, 0.04, 0.33}
\definecolor{pastelred}{rgb}{1.0, 0.41, 0.38}
\definecolor{darksalmon}{rgb}{0.91, 0.59, 0.48}
\definecolor{salmon}{rgb}{1.0, 0.55, 0.41}
\definecolor{lightsalmonpink}{rgb}{1.0, 0.6, 0.6}
\definecolor{candypink}{rgb}{0.89, 0.44, 0.48}

\begin{wrapfigure}{r}{0.53\textwidth}
    \vspace{-3\baselineskip}
    \hspace{-1.0em}

    \resizebox{0.5\textwidth}{!}{
    \begin{tikzpicture}
    \begin{axis}[
        height=5.5cm,
        width=11cm,
        ybar=2pt,
        bar width=14pt,
    	ylabel={\large Accuracy \%},
    	enlargelimits=0.1,
    	legend style={at={(0.,1.05)}, anchor=north west},
    	xtick=data,
        xticklabels={\large 1-shot, \large 2-shot, \large 5-shot, \large 1\%, \large KNN, \large Linear},
        y axis line style={opacity=0},
        font=\large,
    ]
    \addplot[steelblue, fill=lightskyblue, bar shift=-8.5pt]
    	coordinates {(1, 47.5) (2,57.3) (3,65.4) (4,70.3) (5,77.4) (6,80.1)};
    \addplot[candypink, fill=lightsalmonpink!75!white, bar shift=8.5pt]
    	coordinates {(1, 55.0) (2,63.4) (3,69.5) (4,73.4) (5,78.6) (6,81.0)};
     \addplot[candypink, fill=lightsalmonpink, bar shift=8.5pt]
    	coordinates {(1, 49.5) (2,58.6) (3,65.9) (4,70.7) (5,77.1) (6,80.2)};
    
    \addplot[only marks, nodes near coords, nodes near coords style={above}, point meta=explicit symbolic] 
        coordinates {(1, 55.0)[\textcolor{Up}{$\uparrow 7.5$}] (2,63.4)[\textcolor{Up}{$\uparrow 6.1$}] (3,69.5)[\textcolor{Up}{$\uparrow 4.1$}] (4,73.4)[\textcolor{Up}{$\uparrow 3.1$}] (5,78.6)[\textcolor{Up}{$\uparrow 1.2$}] (6,81.0)[\textcolor{Up}{$\uparrow 0.9$}]};
    \legend{\normalsize DINO \vitb/8,\normalsize \ens{\dinop}{\ew} (Ours)}
    \end{axis}
    \end{tikzpicture}
    }

    \vspace{-0.5\baselineskip}
    \caption{
        Our improvements to DINO, including baseline improvements (dark) and ensembling (light).
    }
    \vspace{-1.0\baselineskip}
    \label{fig:teaser}
\end{wrapfigure}
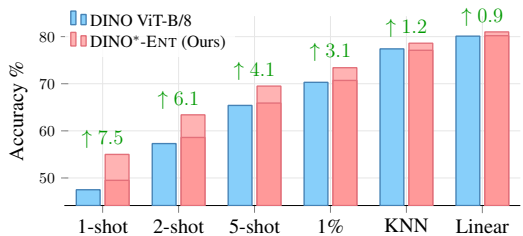
The promise of self-supervised learning (SSL) is to extract information from unlabeled data and leverage this information in downstream tasks \citep{he2020moco,caron2021dino}; \eg, semi-supervised learning \citep{chen2020simclr,chen2020simclrv2}, robust learning \citep{radford2021clip,ruan2022optdom,lee2021compressive}, few-shot learning \citep{assran2022msn}, and supervised learning \citep{tomasev2022pushing}.
These successes have encouraged increasingly advanced SSL techniques
\citep[\eg,][]{grill2020byol,zbontar2021barlowtwins,he2022mae}. 
Perhaps surprisingly however, a simple and otherwise common idea has received limited consideration: ensembling.

Ensembling combines predictions from multiple trained models and has proven effective at improving model accuracy \citep{hansen1990neural,perrone1992networks} and capturing predictive uncertainty in supervised learning \citep{lakshminarayanan2017deepensemble,ovadia2019can}. 
Ensembling in the SSL regime is nuanced, however; since the goal is to learn useful representations from unlabeled data, it is less obvious \emph{where} and \emph{how} to ensemble.
We explore these questions in this work.

We develop an efficient ensemble method tailored for SSL that replicates the non-representation parts (\eg, projection heads) of the SSL model.
\rebedit{In contrast with traditional ``post-training'' ensembling, our ensembles are only used during training to} facilitate the learning of a \emph{single} representation encoder, which yields no extra cost in downstream evaluation.
We further present a family of weighted cross-entropy losses to effectively train the ensembles.
The key component of our losses is the introduction of \emph{data-dependant importance weights} for ensemble members.
We empirically compare different choices from our framework and find that the choice of weighting schemes critically impacts ensemble diversity, and that greater ensemble diversity correlates with improved downstream performance.
Our method is potentially applicable to many SSL methods; we focus on DINO \citep{caron2021dino} and MSN \citep{assran2022msn} to demonstrate its effectiveness.
\Cref{fig:teaser} shows DINO improvements from using our ensembling and weighted cross-entropy loss.

In summary, our core contributions are to:
\begin{itemize}[leftmargin=15pt,itemsep=0pt,topsep=0pt]
\item Develop a downstream-efficient ensemble method suitable for many SSL techniques (\cref{sec:where_ensemble}).
\item Characterize an ensemble loss family of \emph{weighted} cross-entropy objectives (\cref{sec:how_ensemble}).
\item Conduct extensive ablation studies that improve the prior art baselines by up to 6.3 \pp (\cref{sec:exp_heads}).
\item Further improve those baselines with ensembling (\eg, up to 5.5 \pp gain for 1-shot) (\cref{table:compare_ens_merge}).
\end{itemize}


\section{Background}\label{sec:background} 


    


In this section, we frame SSL methods from the perspective of maximum likelihood estimation (MLE) and use this as the notational basis to describe the state-of-the-art clustering-based SSL methods \rebedit{as well as derive their ensembled variants in \cref{sec:method}}.

\paragraph{From Maximum Likelihood to SSL}
Denote unnormalized KL divergence \citep{dikmen2014learning} between non-negative integrable functions $p,q$ by $\KK[p(X),q(X)]=\HH^\times[p(X),q(X)]-\HH[p(X)]$,
where $\HH^\times[p(X),q(X)] = -\int_\X p(x)\log q(x) \dd x + \int_\X q(x)\dd x-1$
is the unnormalized cross-entropy (with $0\log 0=0$) and $\HH[p(X)]=\HH^\times[p(X),p(X)]$. 
These quantities simplify to their usual definitions when $p,q$ are normalized, but critically they enable flexible weighting of distributions \rebedit{for the derivation of our weighted ensemble losses in \cref{sec:how_ensemble}}.

Let $\nu(X, Y)=\nu(X)\nu(Y|X)$ be nature's distribution of
input/target pairs over the space $\X\times \Y$ and $s(Y|\theta,X)$ be a predictive model of target given the input parameterized by $\theta \in \mathcal{T}$.
Supervised maximum likelihood seeks the minimum expected conditional population risk with respect to $\theta$, 
\begin{align}
\label{eq:mle_loss}
\EE_{\nu(X)}\KK[\nu(Y|X),s(Y|\theta,X)] = \EE_{\nu(X)}\HH^\times[\nu(Y|X),s(Y|\theta,X)]-\EE_{\nu(X)}\HH[\nu(Y|X)].
\end{align}

Henceforth omit $\EE_{\nu(X)}\HH[\nu(Y|X)]$ since it is constant in $\theta.$ Since $\nu(X, Y)$ is unknown, a finite sample approximation is often employed.
Denote a size-$n$ \iid training set by  $\mathcal{D}_n=\{x_i\}_{i\in[n]}\sim\nu^{\otimes n}$ and empirical distribution by $\hat \nu(X, Y)=\frac{1}{n}\sum_{x\in\mathcal{D}_n,y\sim\nu(Y|x)}\delta(X-x)\delta(Y-y)$ where $\delta:\reals\to\{0,1\}$ is $1$ when $x=0$ and $0$ otherwise. 
The sample risk is thus
$-\tfrac{1}{n} \sum_{x\in\mathcal{D}_n}\HH^\times[\hat\nu(Y|x),s(Y|\theta,x)].$

In SSL, we interpret $\nu(Y|x)$ as being the oracle teacher under a presumption of how the representations will be evaluated on a downstream task. This assumption is similar to that made in~\citet{arora2019theoretical,nozawa2020pacbayescurl}. 
We also assume $\hat\nu(Y|X)$ is inaccessible and/or unreliable. 
Under this view, some SSL techniques substitute $\hat\nu(Y|x)$ for a weakly learned target or ``teacher'', $t(Y|x).$ 
We don't generally expect $t(Y|x)$ to recover $\nu(Y|x)$; we only assume that an optimal teacher exists and it is $\nu(Y|x)$.
With the teacher providing the targets, the loss becomes
$-\tfrac{1}{n} \sum_{x\in\mathcal{D}_n}\HH^\times[t(Y|x),s(Y|\theta,x)].$

\paragraph{Teacher and student in clustering SSL methods}

Clustering SSL methods such as SWaV \citep{caron2020swav}, DINO \citep{caron2021dino}, and MSN \citep{assran2022msn} employ a student model characterized by proximity between learned codebook entries and a data-dependent code,
\begin{align}
s(Y|\theta,x) &= \softmax\left(\left\{\frac{1}{\tau}\frac{(h_\psi\circ r_\omega)(x)\cdot \mu_y}{\|(h_\psi\circ r_\omega)(x)\|_2\|\mu_y\|_2}:y\in[c]\right\}\right)\label{eq:student}\\
\theta&=\{\omega,\psi,\{\mu_y\}_{y\in[c]}\}\in\mathcal{T},\label{eq:params}
\end{align}
where the encoder $r_\omega:\X\to\Z$ produces the representations used for downstream tasks, and the projection head $h_\psi:\Z \to \reals^d$ and codebook entries $\{\mu_y\}_{y\in\Y}\in\reals^d$ characterize the SSL loss.
\cref{eq:student} can be viewed as ``soft clustering'', where the input is assigned to those centroids that are closer to the projection head's output.
\rebedit{The projection head and codebook are used during training but thrown away for evaluation, which is empirically found vital for downstream tasks \citep{chen2020simclr,chen2020simclrv2}.}
Hyperparameters $\tau\in\reals_{>0},c\in\ints_{>0}$ represent temperature and codebook size.
The teacher is defined as $t(Y|x) = s(Y|\stopgrad(g(\theta)),x)$ where $g:\mathcal{T}\to\mathcal{T}$.
Commonly $g(\theta)$ is an exponential moving average of gradient descent iterates and the teacher uses a lower temperature than the student.

To capture desirable invariances and prevent degeneracy, data augmentation and regularization (\eg, Sinkhorn-Knopp normalization \citep{caron2020swav}, mean entropy maximization \citep{assran2022msn}) are essential.
As these are not directly relevant to our method, we omit them for brevity.

\section{Method}
\label{sec:method}

Ensembling is a technique that combines models to boost performance, and has been especially successful in supervised learning.
We are interested in ensembling methods that carry over this success to SSL approaches. 
However, SSL has key differences, such as throw-away ``projection heads'', from supervised learning that result in a multitude of possibilities for how to ensemble.
With this in mind, we propose first \emph{where} to ensemble, and then \emph{how} to ensemble. 
Those proposals result in \rebedit{an efficient ``peri-training'' ensembling technique specifically tailored for SSL and a family of \emph{weighted} ensemble objectives}; we subsequently suggest different ways to select the weights.

\subsection{Where to ensemble?}
\label{sec:where_ensemble}

\begin{wrapfigure}{r}{0.45\textwidth}
    \vspace{-3.0\baselineskip}
    \hspace{-1.0em}
    \resizebox{0.5\textwidth}{!}{
        \resizebox{\textwidth}{!}{
\begin{tikzpicture}[x=1em,y=1em]
    \node [input] (x) {$x$};
    
    \node[nn, fill=white, shape border rotate=270, minimum size=2em] (f_s) at  ($(x)+(4.5,-5)$) {Student $r_s$};
    \node[nn, shape border rotate=270, minimum size=2em] (f_t) at ($(x)+(4.5,5)$) {Teacher $r_t$};
    
    \node[nn, fill=white,shape border rotate=270, minimum size=2em] (h_s_1) at  ($(f_s)+(6,2)$) {$h_1$};
    \node [input] (vdot_h_s) at  ($(f_s)+(6,0.2)$) {$\vdots$};
    \node[nn, fill=white,shape border rotate=270, minimum size=2em] (h_s_m) at  ($(f_s)+(6,-2)$) {$h_m$};
    \node[nn, shape border rotate=270, minimum size=2em] (h_t_1) at ($(f_t)+(6,+2)$) {$h_1$};
    \node [input] (vdot_h_t) at  ($(f_t)+(6,0.2)$) {$\vdots$};
    \node[nn, shape border rotate=270, minimum size=2em] (h_t_m) at ($(f_t)+(6,-2)$) {$h_m$};
    
    \pic["$s_1$"] (p_s_1) at ($(h_s_1)+(4,0)$) {bars1={0.95}{0.7}{0.45}{0.55}{fill=white}};
    \node [input] (vdot_p_s) at  ($(vdot_h_s)+(3.5,0.2)$) {$\vdots$};
    \pic["$s_m$"] (p_s_m) at ($(h_s_m)+(4,0)$) {bars1={0.6}{0.7}{0.45}{0.8}{fill=white}};
    \pic["$t_1$"] (p_t_1) at ($(h_t_1)+(4,0)$) {bars1={2}{0.5}{0.25}{1}{fill=black!10}};
    \node [input] (vdot_p_t) at  ($(vdot_h_t)+(3.5,0.2)$) {$\vdots$};
    \pic["$t_m$"] (p_t_m) at ($(h_t_m)+(4,0)$) {bars1={1}{0.35}{0.5}{2}{fill=black!10}};
    
    \node [filter,fill=white] (l_m) at ($(f_s)+(14.5,2)$) {$\HH^\times_{mm}$};
    \node [input] (vdot3) at ($(f_s)+(14.5,5.2)$) {$\vdots$};
    \node [filter,fill=white] (l_1) at ($(f_t)+(14.5,-2)$) {$\HH^\times_{11}$};
    
    \node [op,fill=white] (full_l_m) at ($(f_s)+(17.5,5)$) {$+$};
    \node [output] (output) at ($(f_s)+(19.2,5)$) {};
    
    \coordinate (e_mvg_avg_tail) at ($(f_s)+(0,2)$);
    \coordinate (e_mvg_avg_head) at ($(f_t)+(0,-2)$);
    \coordinate (h_mvg_avg_tail) at ($(f_s)+(6,4)$);
    \coordinate (h_mvg_avg_head) at ($(f_t)+(6,-4)$);
    \coordinate (encoders_btm) at ($(f_s)+(0,-3)$);
    \coordinate (heads_btm) at ($(f_s)+(4,-3)$);
    \coordinate (predictions_btm) at ($(f_s)+(10,-3)$);
    \coordinate (loss_btm) at ($(f_s)+(15.5,-3)$);
    
    \node[input] (encoders) at ($(f_t)+(0,4)$) {Encoders};
    \node[input] (heads) at ($(f_t)+(6,4)$) {Heads};
    \node[input] (predictions) at ($(f_t)+(10,4)$) {Preds.};
    \node[input] (losses) at ($(f_t)+(14.5,4)$) {Losses};
    \node[input] (w_11) at ($(l_1)+(0,3)$) {$w_{11}$};
    \node[input] (w_mm) at ($(l_m)+(0,-3)$) {$w_{mm}$};
    
    \draw[->,rounded corners=2ex, gray] (x) |- node[below,near end]{\color{black}$x''$}(f_s);
    \draw[->,rounded corners=2ex, gray] (x) |- node[above,near end]{\color{black}$x'$}(f_t);
    
    \draw[->, gray] (f_s) .. controls ($(f_s)+(4, 1)$) and ($(h_s_1)+(-2, 0)$) .. (h_s_1);
    \draw[->, gray] (h_s_1) -- (p_s_1-in);
    \draw[->, gray] (f_s) .. controls ($(f_s)+(4, -1)$) and ($(h_s_m)+(-2, 0)$) .. (h_s_m);
    \draw[->, gray] (h_s_m) -- (p_s_m-in);
    \draw[->, gray] (f_t) .. controls ($(f_t)+(4, 1)$) and ($(h_t_1)+(-2, 0)$) .. (h_t_1);
    \draw[->, gray] (h_t_1) -- (p_t_1-in);
    \draw[->, gray] (f_t) .. controls ($(f_t)+(4, -1)$) and ($(h_t_m)+(-2, 0)$) .. (h_t_m);
    \draw[->, gray] (h_t_m) -- (p_t_m-in);
    
    \draw[->, gray] (p_s_1-out) .. controls ($(p_s_1-out)+(1, 0)$) and ($(l_1)+(-3, -1)$) .. (l_1.190);
    \draw[->, gray] (p_t_1-out)  .. controls ($(p_t_1-out)+(1, 0)$) and ($(l_1)+(-3, 1)$) ..  node[near end, gray, strike out, draw,-]{} (l_1.170);
    
    \draw[->, gray] (p_s_m-out) .. controls ($(p_s_m-out)+(1, 0)$) and ($(l_m)+(-3, -1)$) ..  (l_m.190);
    \draw[->, gray] (p_t_m-out) .. controls ($(p_t_m-out)+(1, 0)$) and ($(l_m)+(-3, 1)$) .. node[near end, gray, strike out, draw,-]{} (l_m.170);
    
    \draw[->, gray] (l_1)  .. controls ($(l_1)+(2, 0)$) and ($(full_l_m)+(-2, 1)$) ..  (full_l_m);
    \draw[->, gray] (l_m) .. controls ($(l_m)+(2.1, 0)$) and ($(full_l_m)+(-1.8, -1)$) ..  (full_l_m);
    
    \draw[->, gray] (full_l_m) -- (output);
    \draw[->, gray] (w_11) -- node[near start, gray, strike out, draw,-]{} (l_1);
    \draw[->, gray] (w_mm) -- node[near start, gray, strike out, draw,-]{} (l_m);
    
    \draw[->, densely dotted, gray] (e_mvg_avg_tail) -- node[right,midway,align=left]{\color{black}moving\\\color{black}average}(e_mvg_avg_head);
    \draw[->, densely dotted, gray] (h_mvg_avg_tail) -- node[right,midway,align=left]{}(h_mvg_avg_head);
    
    \begin{pgfonlayer}{background}
      \node[fill=cyan,opacity=.08, rounded corners=1ex, fit=(encoders)(f_s)(f_t)(encoders_btm), inner xsep=1.0ex, inner ysep=1.0ex, minimum height=10ex] (encoders_back) {};
    
      \node[fill=cyan,opacity=.08, rounded corners=1ex, fit=(heads)(heads_btm)(h_s_1)(h_s_m)(h_t_1)(h_t_m), inner xsep=0.5ex, inner ysep=1.0ex, minimum height=10ex] (heads_back) {};
    
      \node[fill=cyan,opacity=.08, rounded corners=1ex, fit=(predictions)(p_s_1-in)(p_t_m-out)(predictions_btm), inner xsep=0.25ex, inner ysep=1.0ex, minimum height=10ex] (predictions_back) {};
    
      \node[fill=cyan,opacity=.08, rounded corners=1ex, fit=(losses)(l_1)(l_m)(loss_btm), inner xsep=0.25ex, inner ysep=1.0ex, minimum height=10ex] (losses_back) {};
    
    \end{pgfonlayer}
  \end{tikzpicture}
}
    }
    \caption{
        \textbf{Overview of $(h_\psi,\mu)$-ensemble.}
        Two augmented inputs are encoded by the teacher/student into representations, and then processed by an ensemble of heads.
        The loss for each head is weighted and summed into the final loss.
        Strike-through edges indicate stop-gradients.
        See \cref{appx:pseudocode} for pseudocode.
    }
    \vspace{-2.0\baselineskip}
    \label{fig:approach}
\end{wrapfigure}
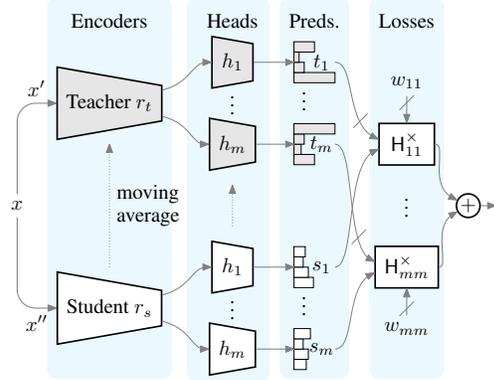
Denote the teacher/student ensembles by $\{t_i(Y|x)\}_{i \in [m]}$ and $\{s(Y|\theta_j,x)\}_{j \in [m]}$ and define each as in \Cref{sec:background}; parameters $\theta=\set{\theta_j}_{j \in [m]}\in\mathcal{T}^m$ are independently initialized, all students use one temperature and all teachers another. 
We asymmetrically denote $t_i(Y|x)$ and $s(Y|\theta_j,x)$ to emphasize that teachers' gradients are zero and that the students are distinct solely by way of $\theta_i\ne\theta_j.$
Studying heterogeneous architectures and/or different teacher parameterizations is left for future work.

Recall that $\theta_j$ parameterizes the encoder, projection head, and codebook parameters: $\theta_j=(\omega_j,\psi_j,\{\mu_{jy}\}_{y\in\Y})$. 
We further restrict $\mathcal{T}^m$ such that $\omega_i=\omega_j$, i.e., 
we limit our consideration to ensembles of projection heads $h_{\psi_j}$ and/or codebooks $\mu_j$ but not encoders $r_{\omega_j}$. 
\rebedit{This choice makes our ensemble method inherently different from traditional supervised ensembling or encoder $r_\omega$ ensembling: the ensembled parts are not used for evaluation but for improving the learning of non-ensembled representation encoder during training, thus it requires no change of downstream evaluation or computational cost.}
Ensembling of $r_\omega$ is left for future work.


\subsection{How to ensemble?}
\label{sec:how_ensemble}

We would like to extend the loss to support an ensemble of teacher/student pairs \rebedit{while respecting the MLE intuition of the loss as in \cref{sec:background}}.
Additionally, we want to facilitate data-dependent importance weights, thus enabling preferential treatment of some teacher/student pairs.
We therefore propose a weighted average (unnormalized) cross-entropy loss,
\begin{align}
\mathcal{L}_n(\theta)&= \frac{1}{n}\sum_{x\in\mathcal{D}_n} \sum_{i,j\in[m]}  \HH^{\times}[w_{ijY} \odot t_i(Y|x),s(Y|\theta_j,x)]\label{eq:ensemble_loss}\\
\text{where}\quad w_{ijy}&=\softmax\left(\left\{\tfrac{1}{\gamma}f_{ijy}(\stopgrad(\theta),x) :i,j\in[m]\right\}\right).\label{eq:ensemble_weight}
\end{align}
The notation $w_{ijY}\odot t_i(Y|x)$ denotes a Hadamard product; \ie, the product of event-specific weights and probabilities for each $y\in\Y.$
The hyperparameter $\gamma$ is the temperature. 
The function $f_{ijy}$ is defined for brevity and discussed in the following section.

This objective admits generality and flexibility for introducing various weighting schemes, as it supports potential interactions between all teacher/student pairs and allows the weights to be both model- and data-dependent.
Up to a constant independent of $\theta$, it is an importance weighted average of (unnormalized) KL divergences between each teacher and each student; \ie, a mixture of MLE-like objectives. 
We stop the gradient of $w_{ijy}$ to $\theta$ in order to keep the overall gradient a weighted average of students' log-likelihood gradients, similar to \cref{eq:mle_loss}.
We also normalize the weights such that each data point equally contributes to the loss.


\subsection{How to weight?}\label{sec:weight_ensemble}

In this section, we present several instantiations of our losses with different weighting schemes. 
We empirically show in \cref{sec:experiments} that the particular choice of weighting scheme is critical for the representation performance and the induced diversity of $(h_\psi,\mu)$-ensembles.
For simplicity we assume $\gamma = 1$ in this section.
We indicate with $\iff$ that a loss has the same $\arg\min$ as \Cref{eq:ensemble_loss}.
For additional analysis and discussion, see \Cref{app:analysis}.

\paragraph{Uniform weighting (\outside)}
\label{loss:unif}
The simplest strategy is to treat different teacher/student pairs independently and average each with uniform weighting; \ie,
\begin{align}
    f_{ijy}=\log \delta(i-j) \iff \mathcal{L}_n^\text{\outside}(\theta) = \frac{1}{n} \sum_{x\in\mathcal{D}_n} \frac{1}{m}\sum_{i\in[m]} 
    \HH^\times[t_i(Y|x), s(Y|\theta_i,x)] \label{eq:outside}
\end{align}
This strategy introduces \textit{uniform} weights $w_i = \frac{1}{m}$ over ensemble elements.
The role of $\log\delta(i-j)$ (here and elsewhere) is to sub-select corresponding teacher/student pairs rather than all $m^2$ pairs.

\paragraph{Probability weighting (\inside)}
\label{loss:prob}
An alternative to using the average cross-entropy loss (\outside) is to compute the cross-entropy loss of the average predictions whose gradient is weighted by $w_{ijy}$ (see \Cref{appx:derivations}).
At $\gamma=1$, those gradient weights simplify into an average over the student probabilities:
\begin{align}
    f_{ijy}=\log s(y|\theta_j,x) \iff \mathcal{L}^\text{\inside}_n(\theta) = \frac{1}{n} \sum_{x\in\mathcal{D}_n}  
    \HH^\times\left[\frac{1}{m}\sum_{i\in[m]}t_i(Y|x), \frac{1}{m}\sum_{j\in[m]} s(Y|\theta_j,x)\right]
    \label{eq:inside}
\end{align}
Averaging the predictive distributions introduces correspondence between codes from different heads; thus different heads are no longer independent but instead \emph{cooperate} to match the student to the teachers.
The loss favors student heads with more confident predictions (\ie, larger $s(y|\theta_j,x)$).
Further motivation for averaging predictions comes from multi-sample losses studied in \citet{morningstar2022pacm}.
Note that the joint convexity of (unnormalized) KL divergence implies that this loss is upper bounded by the \outside loss up to some constant in $\theta$ (see \Cref{app:analysis}).

Although the \inside strategy favors confident student predictions, the weights change as a function of $y\in\Y$. 
This may be in conflict with our intuition that SSL is like maximum likelihood (\Cref{sec:background}), since under that view, the teacher is responsible for weighting outcomes.

\paragraph{Entropy weighting (\ew)}
\label{loss:ent}
Another way to favor heads with more confident predictions is to directly weight by their predictive entropies; \ie,
\begin{align}
    f_{ijy}&=-\HH[t_i(Y|x)]+\log \delta(i-j)
     \iff\\
    \mathcal{L}^\text{\ew}_n(\theta) &= \frac{1}{n} \sum_{x\in\mathcal{D}_n}  
    \sum_{i\in[m]} \softmax_i(\{-\tfrac{1}{\gamma}\HH[t_{i'}(Y|x)]:i'\in[m]\}) \HH^\times\left[ t_i(Y|x), s(Y|\theta_i,x)\right]
    \label{eq:ent_weight}
\end{align}
where the weight $w_i=\softmax_i(\{-\frac{1}{\gamma}\HH[t_{i'}(Y|x)]:i'\in[m]\})$ is inversely correlated with the entropy of teacher predictions.
In other words, the head whose teacher has a lower entropy (i.e., higher confidence about its prediction) is given a larger importance weight for learning the representation.
Like \inside, this strategy encourages ``data specialists'' by emphasizing strongly opinionated teacher heads for different inputs.
Like \outside, different heads are treated more independent (than \inside), since interaction between different heads is  introduced only through the weight computation.
By preferring low-entropy teachers we also favor low variance teachers; this aligns with the intuition that using a lower-variance teacher benefits representation quality \citep{wang2022asym}.

\paragraph{Countless other weighting schemes} 
It is impossible to fully explore the space of weightings; the following might also be interesting to study in detail but were omitted due to resource constraints.
\begin{align}
f_{ijy} &=0                                    && \text{(Favors all pairs of teachers/students equally)} \label{eq:fully_uniform} \\
f_{ijy} &= \log t_i(y|x)                       && \text{(Favors opinionated teachers)} \label{eq:opinionated_teacher} \\
f_{ijy} &= -\HH[s(Y|\theta_j,x)]              && \text{(Favors low-entropy students)} \label{eq:low_ent_student} \\
f_{ijy} &= \KK[t_i(Y|x),s(Y|\theta_j,x)]        && \text{(Favors disagreeing teacher/student pairs)} \label{eq:disagree_pairs} \\
f_{ijy} &= -\tfrac{1}{2}\log(\Var_{t_i(Y|x)}[Y] + \epsilon)  && \text{(Favors low variance teachers; e.g., $\epsilon=\tfrac{1}{12}$)}  \label{eq:low_var_teacher}
\end{align}
Note that ``aligned'' versions of all schemes are possible by using $f_{ijy}+\log\delta(i-j).$
We did early experiments exploring \cref{eq:opinionated_teacher,eq:low_ent_student}, but the results were inferior and are largely omitted below.



\section{Related Work}

\paragraph{Self-supervised learning}
Recent work on self-supervised learning (SSL) focuses on discriminative or generative approaches.
Most discriminative approaches seek to learn augmentation-invariant representations by enforcing the similarity between augmented pairs of the same image while utilizing different techniques to avoid collapse.
Contrastive methods \citep{chen2020simclr,he2020moco,wu2018instdisc,hjelm2018infomax,bachman2019infomax,tian2020cmc}
use a large number of negative samples with a noise-contrastive objective \citep{gutmann2010nce,oord2018cpc}.
A large body of followup work eliminates the necessity of explicit negative samples with various techniques, including clustering assignment constraints \citep{caron2018deepcluster,caron2020swav,caron2021dino,asano2019selflabel}, bootstrapping \citep{grill2020byol} or self-distillation \citep{caron2021dino} inspired by mean teacher \citep{tarvainen2017meanteacher}, asymmetric architecture design \citep{grill2020byol,chen2021simsiam}, or redundancy reduction \citep{zbontar2021barlowtwins,bardes2021vicreg}.
Recent generative approaches that use masked image modeling as the pretraining task \citep{dosovitskiy2020image,bao2021beit,he2022mae,zhou2022ibot,xie2022simmim} have achieved competitive finetuning performance. 
Our method may be applicable to all of the above methods that have some sort of ``projection head'', such as most of the discriminative approaches.

\paragraph{Ensemble methods}
Ensembling has been extensively studied for improving model performance \citep{hansen1990neural,perrone1992networks,dietterich2000ensemble} and uncertainty estimation \citep{lakshminarayanan2017deepensemble,ovadia2019can} in supervised learning and semi-supervised learning \citep{laine2016temporal}.
A major research direction is to train efficient ensembles with partial parameter sharing \citep{lee2015multihead,wen2020batchensemble,dusenberry2020efficient,havasi2020mimo} or intermediate checkpointing \citep{huang2017snapshot,garipov2018loss}.
Our method also shares the encoder parameters across ensembles, which is closely related to multi-headed networks \citep{lee2015multihead,tran2020hydra}.
Ensemble methods for SSL are less explored. 
Some recent work studies ensembles of supervised models adapted from pretrained SSL models. 
\citet{gontijo2022no} conduct an empirical study of ensembles adapted from different SSL models and find that higher divergence in SSL methods leads to less correlated errors and better performance.
\citet{wortsman2022modelsoup} ensemble multiple finetuned models adapted from the same SSL model by averaging their weights, which boosts the performance without any inference cost. 
Our method differs from them in that it (1) applies to the SSL training stage to directly improve representation quality, \rebedit{rather than aggregates multiple models in the post-training/finetuning} stage; (2) introduces little training cost and no evaluation cost; and \rebedit{(3) is complementary to these post-training/finetuning ensembling methods.}

\section{Experiments}
\label{sec:experiments}

We carefully study the impact of $(h_\psi,\mu)$-ensembles and our selected weighted ensemble losses (\outside, \inside, and \ew) on smaller DINO models in \Cref{sec:exp_heads}.
Using what we learned in those experiments, in \Cref{sec:sota} we present new state-of-the-art results on ImageNet-1K on various metrics for multiple model sizes by ensembling both DINO- and MSN-based models.
Finally, we explore ensemble evaluations in the transfer learning setting in \Cref{sec:transfer}.
Additional experimental details and results are in \cref{appx:exp_details} and \cref{appx:additional_results}, respectively.

\paragraph{Experimental setup}
We assessed the effectiveness of our method with two SSL methods: DINO \citep{caron2021dino} and MSN \citep{assran2022msn}. 
In order to ensure that we are comparing against strong baselines, we consider three different classes of baselines: 
\textbf{(1)} evaluation numbers reported in the original works (\citet{caron2021dino}, \citet{assran2022msn}, and \citet{zhou2022ibot} for an additional baseline iBOT);
\textbf{(2)} evaluation of our implementation using the hyperparameters reported in the original works (DINO only, for space reasons) to validate our implementation;
and \textbf{(3)} evaluation of our implementation using the best hyperparameters that we found by tuning the baselines (DINO and MSN) for fair comparisons.
In almost all models and evaluations, our retuned baselines give non-trivial performance improvements on top of previously reported numbers.
These type \textbf{(3)} baselines we label \textbf{\dinop} and \textbf{\msnp}, and we use them as the base models for our experiments with $(h_\psi,\mu)$-ensembles and weighted ensemble losses.
\Cref{appx:impl_details} describes the details for getting such strong performance for \dinop and \msnp.
In particular, we find that the projection head has a crucial impact on label efficiency of representations and using a smaller head (3-layer MLP with hidden size 1024) significantly improves few-shot evaluation performance (see \cref{appx:study_proj_head}).

\paragraph{Evaluation metrics}
We compared models trained with and without our $(h_\psi,\mu)$-ensembles by measuring various evaluation metrics on ImageNet-1K \citep{deng2009imagenet}.
The evaluation metrics reflect the \emph{decodability} and the \emph{label efficiency} of learned representations. 
We measured the \emph{decodability} with respect to both the linear classifier following the common linear evaluation protocol and the $k$-NN classifier following \citet{caron2021dino}.
We measured the \emph{label efficiency} by evaluating the linear evaluation performance in few-shot settings, including 1\% ($\sim$13-shots) labeled data evaluation \citep{chen2020simclr} and 1-/2-/5-shot evaluations \citep{assran2022msn}. 
All evaluations used frozen representations of the teacher encoder -- we did not fine tune the models.
See \cref{appx:eval_details} for details.

\subsection{Empirical study of $(h_\psi,\mu)$-ensembles}
\label{sec:exp_heads}

\begin{figure}[tb]
    \centering
    \begin{subfigure}[b]{0.325\linewidth}
            \includegraphics[width=\linewidth, trim={0 0 0 0}, clip]{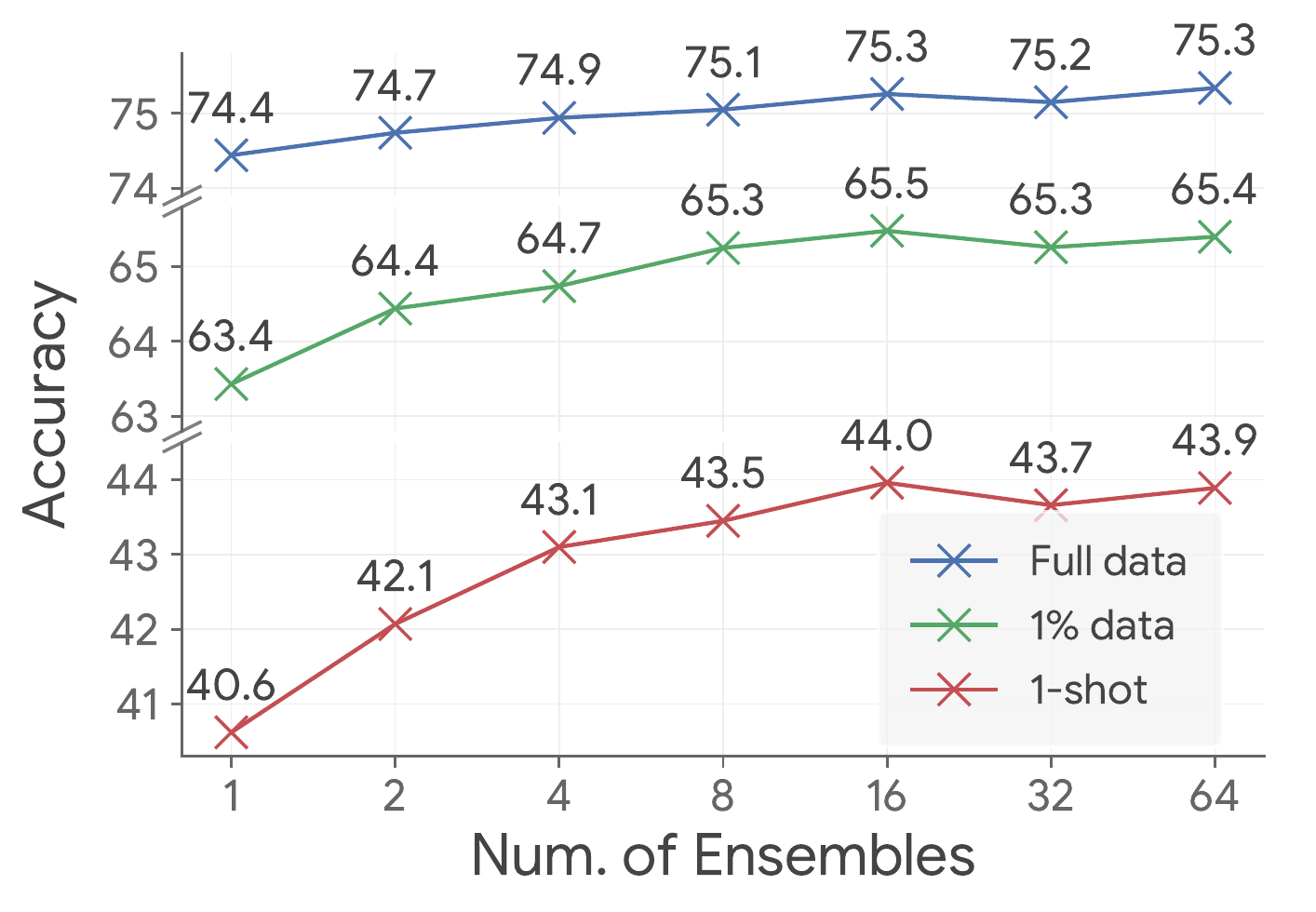}
            \caption{Scaling of $(h_\psi,\mu)$-ensembles.}
            \label{fig:scaling_num_ens_head}
    \end{subfigure}
    \begin{subfigure}[b]{0.325\linewidth}
            \includegraphics[width=\linewidth, trim={0 0 0 0}, clip]{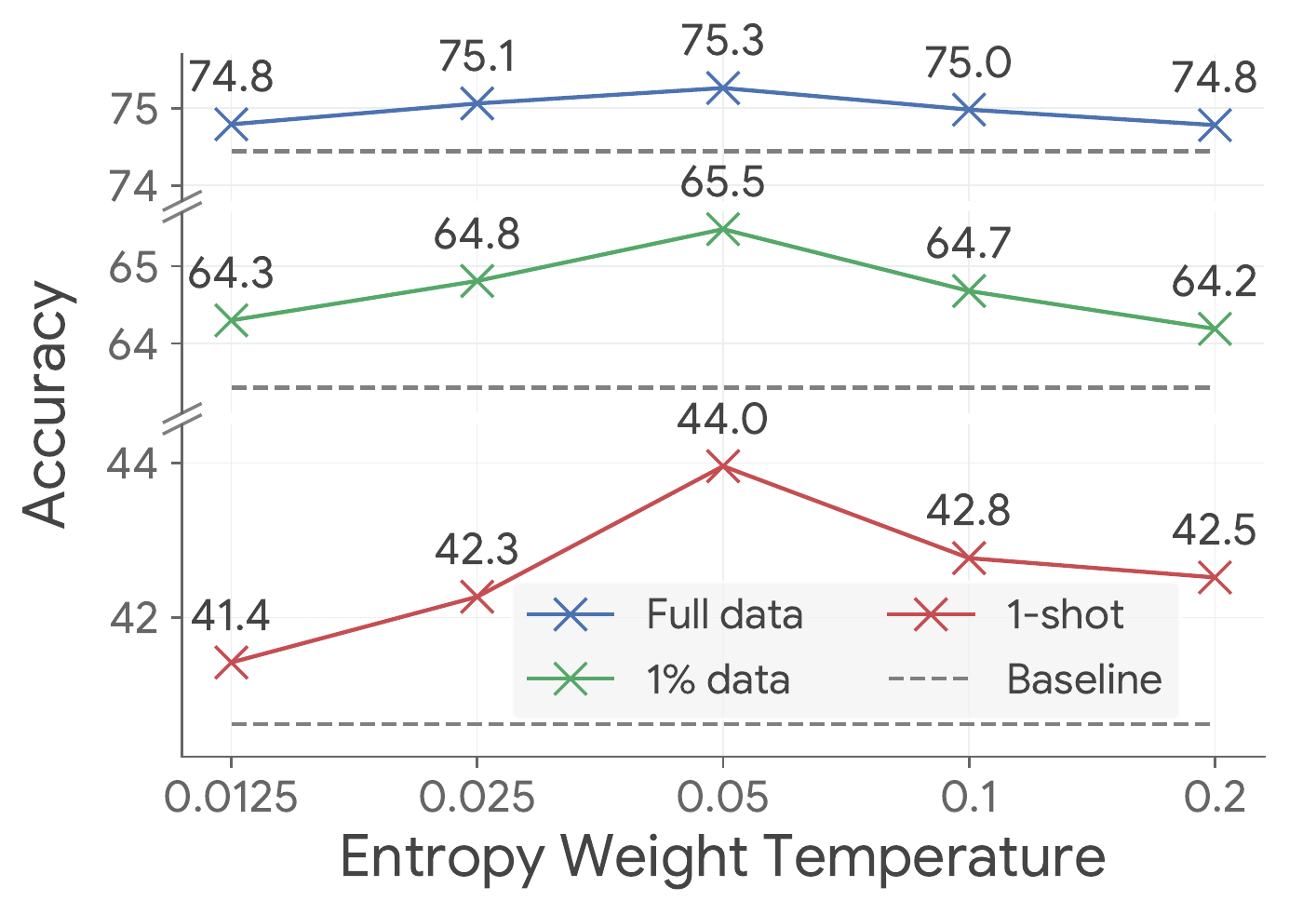}
            \caption{Effect of \ew~temperature $\enttemp$.}
            \label{fig:study_ent_temp}
    \end{subfigure}
    \begin{subfigure}[b]{0.325\linewidth}
            \includegraphics[width=\linewidth, trim={0 0 0 0}, clip]{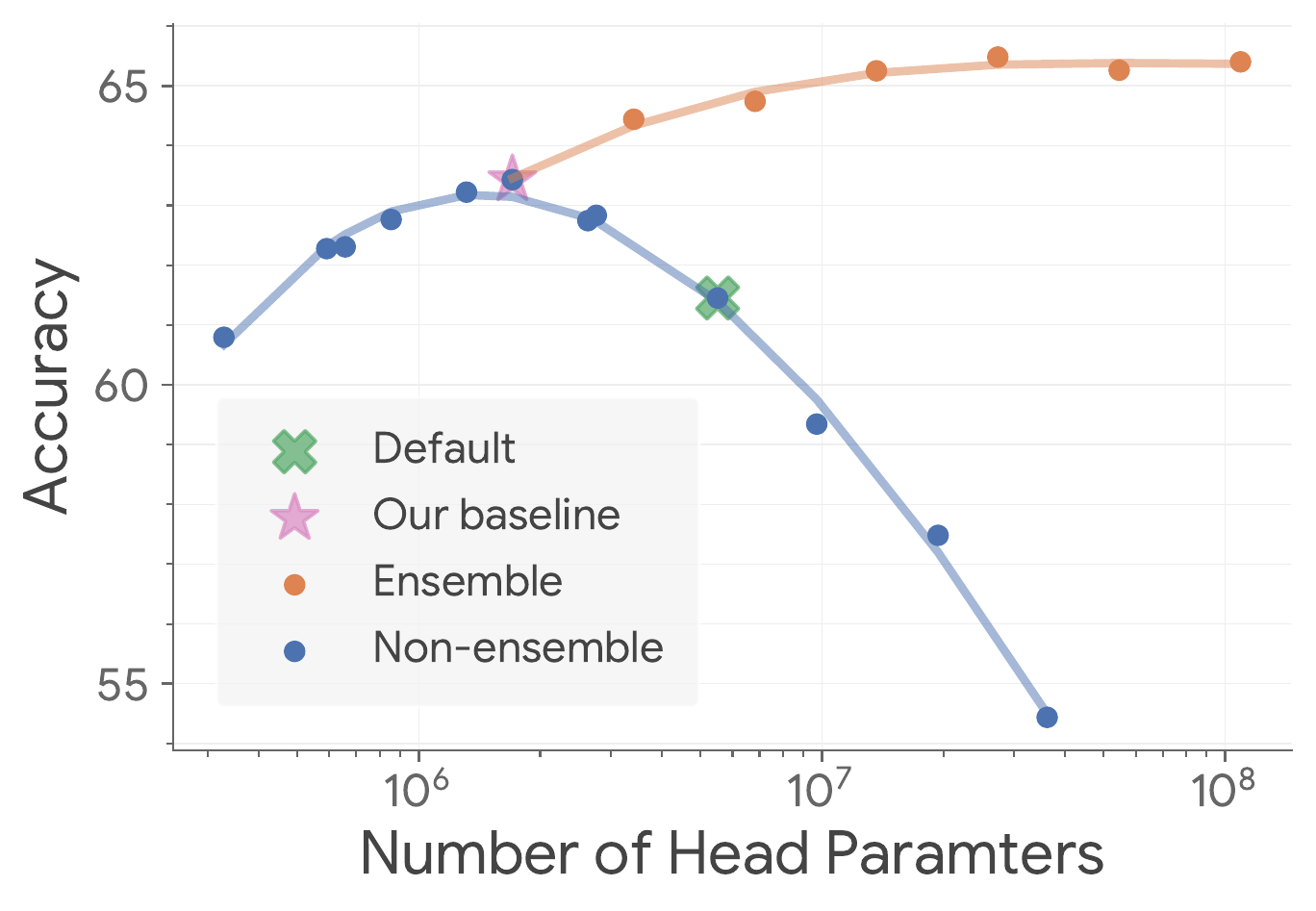}
            \caption{Comparing different heads.}
            \label{fig:compare_head_main}
    \end{subfigure}
    \caption{
        \textbf{Empirical study of $(h_\psi,\mu)$-ensembles.}
        (a) The gains of $(h_\psi,\mu)$-ensembles start to diminish above 16 heads.
        (b) The temperature for entropy weighting has a larger impact on few-shot performance. 
        16 heads are used and $\enttemp$ is scaled by $\log(\numcode)$. 
        (c) Our $(h_\psi,\mu)$-ensembles outperform all non-ensembled baselines when controlling for number of parameters.
        A too powerful non-ensembled projection head significantly harms accuracy. 
        1\%-data evaluation is shown.
        Also see \cref{fig:compare_head_all}.
    }
    \label{fig:compare_head}
    \vspace{-1\baselineskip}
\end{figure}

\begin{wrapfigure}{r}{.35\textwidth}
    \vspace{-1.0\baselineskip}
    \includegraphics[width=0.95\linewidth, trim={0.1in 0.1in 0.1in 0.1in},
    clip]{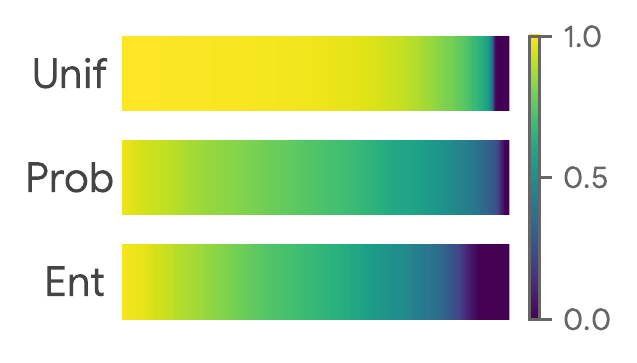}
    \caption{
        \textbf{Visualization of code similarity.} 
        \ew learns the most diverse $(h_\psi,\mu)$-ensembles reflected by the fastest decay of similarity scores between aligned codes in different heads.
        \outside has low diversity between heads.
    }
    \vspace{-\baselineskip}
    \label{fig:vis_ens_head_diversity_decay}
\end{wrapfigure}

\Cref{table:compare_ens_ablation} compares different strategies for \textbf{where} and \textbf{how to ensemble}.
\Cref{fig:vis_ens_head_diversity_decay} compares the impact of the weighted ensemble loss on \textbf{$(h_\psi,\mu)$-ensemble diversity}.
\Cref{fig:compare_head} shows the effect of \textbf{increasing the number of ensembles}, \textbf{adjusting the temperature $\gamma$}, and \textbf{increasing baseline projection head parameters}.
In these experiments, we used \dinop with \vits/16  trained for 300 epochs as the base model.
We compared different ensemble methods applied to the base model with $\numens=16$ heads which we found to work the best.
For the \ew strategy in \cref{table:compare_ens_ablation}, the entropy weighting temperature $\enttemp$ is set to $0.05 \times \log(\numcode)$ by default which is selected from $\set{0.0125, 0.025, 0.05, 0.1, 0.2} \times \log(\numcode)$, where the scale $\log(\numcode)$ gives the maximum entropy of the codebook size $\numcode$.
For \inside, we keep $\gamma=1$.

\begin{table}[t]
\caption{
    \textbf{Comparison of different ensemble strategies.} 
    \ew and \inside significantly improve over the non-ensembled baseline, while \outside leads to no gains.
    Ensembling both the projection head and the codebook works the best.
    All models are \dinop \vits/16 trained for 300 epochs.
    Averages and standard deviations are over 3 initialization seeds.
    The linear evaluation results \rebedit{on ImageNet-1K} with different amounts of labeled data are reported here (see \cref{table:compare_ens_ablation_full} in \cref{appx:study_ens_head} for all metrics).
}
\label{table:compare_ens_ablation}
\vspace{-0.5\baselineskip}
\begin{center}
\small
\adjustbox{max width=\textwidth}{
\begin{tabular}{lccllll}
\toprule
\multirow{2}{*}{\textbf{How}} & \multicolumn{2}{c}{\textbf{Where}} & \multicolumn{4}{c}{\textbf{\# of Labels Per Class}} \\
 \cmidrule(lr){2-3}\cmidrule(lr){4-7}
  & Proj.\ $h_\psi$ &  Code.\ $\mu$ & \multicolumn{1}{c}{1\quad\quad\quad} & \multicolumn{1}{c}{5\quad\quad\quad} & \multicolumn{1}{c}{$\sim$13 (1\%)\quad\quad}  & \multicolumn{1}{c}{Full\quad\quad~} \\ 
 \midrule
 Base & & & 40.6 $\pm$ 0.2 & 57.9 $\pm$ 0.3 & 63.4 $\pm$ 0.2 & 74.4 $\pm$ 0.1\\
 \midrule 
\outside & \checkmark &  \checkmark & 40.4 $\pm$ 0.4 & 57.6 $\pm$ 0.3 &  63.3 $\pm$ 0.3 & 74.5 $\pm$ 0.2\\
\midrule
\inside & \checkmark &   &  39.8 $\pm$ 0.5 \textcolor{Down}{$\downarrow 0.9$} & 57.4 $\pm$ 0.4 \textcolor{Down}{$\downarrow 0.5$} & 63.0 $\pm$ 0.4 \textcolor{Down}{$\downarrow 0.4$}  & 74.8 $\pm$ 0.1 \textcolor{Up}{$\uparrow 0.4$} \\
\inside & \checkmark &  \checkmark & 41.9 $\pm$ 0.3 \textcolor{Up}{$\uparrow 1.3$} & 59.6 $\pm$ 0.4 \textcolor{Up}{$\uparrow 1.7$} & 65.1 $\pm$ 0.3 \textcolor{Up}{$\uparrow 1.7$} & \textbf{75.4 $\pm$ 0.1} \textcolor{Up}{$\uparrow 1.0$} \\
\midrule 
\ew-\textsc{St} & \checkmark &  \checkmark & 40.0 $\pm$ 0.5 \textcolor{Down}{$\downarrow 0.6$} & 57.3 $\pm$ 0.5 \textcolor{Down}{$\downarrow 0.6$} & 62.7 $\pm$ 0.5 \textcolor{Down}{$\downarrow 0.7$} & 74.0 $\pm$ 0.4 \textcolor{Down}{$\downarrow 0.4$} \\
\midrule
\ew &  &  \checkmark &  40.8 $\pm$ 0.4 & 58.0 $\pm$ 0.4 & 63.5 $\pm$ 0.4 & 74.5 $\pm$ 0.3 \\
\ew & \checkmark &   & 43.0 $\pm$ 0.6 
\textcolor{Up}{$\uparrow 2.4$} & 59.7 $\pm$ 0.7 \textcolor{Up}{$\uparrow 1.8$} & 64.8 $\pm$ 0.5 \textcolor{Up}{$\uparrow 1.4$} & 75.1 $\pm$ 0.4 \textcolor{Up}{$\uparrow 0.7$}  \\
\ew & \checkmark &  \checkmark & \textbf{44.0 $\pm$ 0.2} \textcolor{Up}{$\uparrow 3.4$} & \textbf{60.5 $\pm$ 0.3} \textcolor{Up}{$\uparrow 2.6$} & \textbf{65.5 $\pm$ 0.1} \textcolor{Up}{$\uparrow 2.2$} & \textbf{75.3 $\pm$ 0.1} \textcolor{Up}{$\uparrow 0.9$}  \\
\bottomrule
\end{tabular}
}
\end{center}
\vspace{-1.5\baselineskip}
\end{table}

\paragraph{Where to ensemble}
We study the \textbf{where} question by ensembling either the projection head $h_\psi$, the codebook $\mu$, or both with the \ew and the \inside ensemble strategies, as shown in \cref{table:compare_ens_ablation}.
We find that ensembling both $h_\psi$ and $\mu$ provides the largest gains for both losses, probably due to the increased flexibility for learning a diverse ensemble.
Interestingly, only ensembling $h_\psi$ also works well for the \ew strategy. 

\paragraph{How to ensemble}
We study the \textbf{how} question by considering four different loss variants: \outside, \inside, \ew, and the variant of \ew with student entropy weighting.
We find that when we ensemble both the projection head $h_\psi$ and the codebook $\mu$, the \ew ensemble strategy leads to the most significant gains (\eg, 3.4 \pp gains for 1-shot and 0.9 \pp gains for full-data).
The \inside strategy also consistently improves the performance with a slightly larger gain (1 \pp) in full-data evaluation.
In contrast, we see no gains for the \outside strategy over the baseline.
We also study a variant of \ew that uses the student entropy (\ie, \Cref{eq:low_ent_student} with the $\log \delta(i-j)$ term) for the importance weights (denoted as \ew-\textsc{St}). \ew-\textsc{St} performs much worse than \ew and is even worse than the baseline.
We conjecture that this is because the student predictions typically have a larger variance than teacher predictions \citep{wang2022asym} especially when multi-crop augmentation \citep{caron2020swav,caron2021dino} is applied to the student.
Similar experiments on \Cref{eq:opinionated_teacher} and/or $\gamma=0$ variants of \inside also resulted in inferior performance (see \cref{table:compare_ens_prob_variant}).


\paragraph{Analysis of $(h_\psi,\mu)$-ensemble diversity}

The previous experiments showed that the choice of ensemble weighting strategy has a large impact on performance.
We hypothesize that this choice substantially impacts the diversity of the codebook ensembles.
Since the codes in different heads may not be aligned, we align them by the similarity of their code assignment probabilities across different input images, which measures how the codes are effectively used to `cluster' the data.
See \cref{appx:vis_head_diversity} for detailed explanations and results.
In \cref{fig:vis_ens_head_diversity_decay}, we visualize the decay patterns of the similarity score between aligned codes (1.0 means the most similar) in a random pair of heads for each weighting strategy.
\ew decays the fastest and \outside decays the slowest, indicating that \ew learns the most diverse codebooks while \outside is least diverse.
This shows a positive correlation between the diversity of $(h_\psi,\mu)$-ensembles and the empirical performance of the ensemble strategies from \Cref{table:compare_ens_ablation}.
Finally, for \outside, we find that different heads tend to learn the same semantic mappings even when randomly initialized; \ie, the code assignments in different heads become homogeneous up to permutation.
See \cref{fig:vis_ens_head_diversity} for a visualization.



\paragraph{Number of $(h_\psi,\mu)$-ensembles}
We study the effect of increasing the number of $(h_\psi,\mu)$-ensembles $\numens$ for \ew in \cref{fig:scaling_num_ens_head}.
Having more $(h_\psi,\mu)$-ensembles boosts the performance until $\numens = 16$.
Interestingly, using as few as $\numens=2$ heads already significantly improves over the baseline.

\paragraph{Effect of \ew temperature $\enttemp$} 
\cref{fig:study_ent_temp} studies the effect of entropy weighting temperature $\enttemp$ for different evaluation metrics.
We observe that $\enttemp$ has a relatively larger impact on few-shot evaluation performance. 
$\enttemp$ should be neither too high nor too low: a high temperature leads to under-specialization (\ie. less diversity) of heads similar to \outside ($\enttemp \to \infty$) and a low temperature may otherwise lead to over-specialization (\ie, only a single head is used for each input).

\begin{table}[t]
\vspace{-1.5\baselineskip}
\caption{
    \textbf{Effectiveness of ensemble heads} for \dinop/\msnp with different ViT models. 
    Our ensemble heads consistently improve all downstream evaluation metrics \rebedit{on ImageNet-1K} and achieve a new state-of-the-art for few-shot evaluations.
    For \vits/16, we report linear evaluation results probed from the last layer (left) and from the last 4 layers (right, following DINO).
    $^\dagger$We evaluated the few-shot settings using DINO's publicly-available pretrained weights in the cases those results were not reported in \citet{caron2021dino}.
    $^\ddagger$MSN \vitb/16 and \vitb/8 are both trained for 600 epochs in \citet{assran2022msn}, whereas our models are trained for only 400, 300 epochs, respectively.
    For each architecture, we highlight the best \colorbox{LightCyan}{\!DINO\!} baseline and weighted ensemble in \colorbox{LightCyan}{\!blue\!}.
    For \colorbox{LightYellow}{\!MSN\!}, the corresponding highlights are \colorbox{LightYellow}{\!yellow\!}.
    The best results for each architecture and metric are \textbf{bolded}.
}
\vspace{-1\baselineskip}
\label{table:compare_ens_merge}
\begin{center}
\small
\adjustbox{max width=\textwidth}{
\begin{tabular}{llcclcl}
\toprule
\multirow{2}{*}{\textbf{Method}} & \multicolumn{4}{c}{\textbf{Few-shot}} & \multicolumn{2}{c}{\textbf{Full-data}} \\
 \cmidrule(lr){2-5}\cmidrule(lr){6-7}
 & \multicolumn{1}{c}{1\quad\quad\quad} & 2 & 5 & \multicolumn{1}{c}{$\sim$13 (1\%)} & $k$-NN & \multicolumn{1}{c}{Linear} \\ 
 \midrule
 \emph{\vits/16, 800 epochs}\\
 \midrule
  iBOT & 40.4  $\pm$ 0.5  & 50.8 $\pm$ 0.8  &  59.9 $\pm$ 0.2 & 65.9 & \textbf{75.2} & ~~~-~~ / \textbf{77.9} \\
DINO &  38.9 $\pm$ 0.4 & 48.9 $\pm$ 0.3 & 58.5 $\pm$ 0.1 & 64.5 & 74.5 & 76.1 / 77.0 \\ 
DINO (\repro) &  39.1 $\pm$ 0.3  &  49.1 $\pm$ 0.5  &  58.6 $\pm$ 0.2  & 64.7 & 74.3 & 75.8 / 76.9 \\
\rowcolor{LightCyan!80}
\dinop (\retune) & 44.6 $\pm$ 0.2 & 53.6 $\pm$ 0.3   &  61.1 $\pm$ 0.2  & 66.2 & 74.1 & 75.8 / 76.9 \\
MSN &  47.1 $\pm$ 0.1 & 55.8 $\pm$ 0.6 & 62.8 $\pm$ 0.3 & 67.2 & - & ~~~-~~ / 76.9 \\
\rowcolor{LightYellow}
\msnp (\retune) &  47.4 $\pm$ 0.1  & 56.3 $\pm$ 0.4  &  62.8 $\pm$ 0.2   &  67.1  & 73.3 &  75.6 / 76.6 \\ 
\midrule
\ens[16]{\dinop}{\inside}  &  45.2 $\pm$ 0.4  &  54.9 $\pm$ 0.4  &  62.5 $\pm$ 0.2  & 67.3 & 75.1 &  76.5 / 77.6 \\
\ens[4]{\dinop}{\ew}  & 46.3 $\pm$ 0.1   &  55.5 $\pm$ 0.6  & 63.0 $\pm$ 0.3   & 67.5 & 74.8 &  76.2 / 77.2 \\
\rowcolor{LightCyan}
\ens[16]{\dinop}{\ew} &  47.6 $\pm$ 0.1 \textcolor{Up}{$\mathbf{\uparrow 3.0}$}  &  56.8 $\pm$ 0.5  &  64.0 $\pm$ 0.2  & 68.3 \textcolor{Up}{$\mathbf{\uparrow 2.1}$} & \textbf{75.3} & \textbf{76.8} / \textbf{77.7} \textcolor{Up}{$\mathbf{\uparrow 0.8}$}  \\
\ens[2]{\msnp}{\ew} &  48.8 $\pm$ 0.2  & 57.5 $\pm$ 0.5   &  64.0 $\pm$ 0.2   & 67.9 & 74.6 & 76.0 / 76.9  \\
\rowcolor{LightYellow}
\ens[8]{\msnp}{\ew} &  \textbf{50.1 $\pm$ 0.1} \textcolor{Up}{$\mathbf{\uparrow 2.7}$}  & \textbf{58.9 $\pm$ 0.6}   &  \textbf{65.1 $\pm$ 0.3}   & \textbf{68.7} \textcolor{Up}{$\mathbf{\uparrow 1.6}$} & \textbf{75.2} & 76.4 / 77.4 \textcolor{Up}{$\mathbf{\uparrow 0.8}$}  \\
 \midrule
 \emph{\vitb/16, 400 epochs} \\
 \midrule
 iBOT & 46.1 $\pm$ 0.3 & 56.2 $\pm$ 0.7  & 64.7 $\pm$ 0.3 & 69.7 & \textbf{77.1} & \quad\textbf{79.5} \\
DINO$^\dagger$ & 43.0 $\pm$ 0.2 & 52.7 $\pm$ 0.5 & 61.8 $\pm$ 0.2 & 67.4 & 76.1 & \quad78.2 \\ 
\rowcolor{LightCyan}
 \dinop (\retune) & 49.3 $\pm$ 0.1 & 58.1 $\pm$ 0.5  & 65.0 $\pm$ 0.3  & 69.1 & 76.0 & \quad78.5 \\ 
 MSN$^\ddagger$ & 49.8 $\pm$ 0.2 & 58.9 $\pm$ 0.4 & 65.5 $\pm$ 0.3 & ~~~-~~~ & - & \quad~~~- \\
 \rowcolor{LightYellow}
 \msnp (\retune) & 50.7 $\pm$ 0.1 & 59.2 $\pm$ 0.4 & 65.9 $\pm$ 0.2 & 69.7 & 74.7 & \quad78.1\\
 \midrule
 \rowcolor{LightCyan}
 \ens[16]{\dinop}{\ew} & 52.8 $\pm$ 0.1 \textcolor{Up}{$\mathbf{\uparrow 3.5}$} & 61.5 $\pm$ 0.4  &  67.6 $\pm$ 0.3 & 71.1 \textcolor{Up}{$\mathbf{\uparrow 2.0}$} & \textbf{77.1}  & \quad79.1 \textcolor{Up}{$\mathbf{\uparrow 0.6}$} \\
 \rowcolor{LightYellow}
 \ens[8]{\msnp}{\ew} & \textbf{53.7 $\pm$ 0.2} \textcolor{Up}{$\mathbf{\uparrow 3.0}$} & \textbf{62.4 $\pm$ 0.6} & \textbf{68.3 $\pm$ 0.2} & \textbf{71.5} \textcolor{Up}{$\mathbf{\uparrow 1.8}$} & \textbf{77.2} & \quad78.9 \textcolor{Up}{$\mathbf{\uparrow 0.8}$} \\
 \midrule
 \emph{\vitb/8, 300 epochs} \\
 \midrule
DINO$^\dagger$ & 47.5 $\pm$ 0.2 & 57.3 $\pm$ 0.5 & 65.4 $\pm$ 0.3 & 70.3 & 77.4 & \quad80.1  \\  
\rowcolor{LightCyan}
\dinop (\retune) & 49.5 $\pm$ 0.5 & 58.6 $\pm$ 0.6  & 65.9 $\pm$ 0.3 & 70.7 & 77.1 & \quad80.2\\
 MSN$^\ddagger$ & 55.1 $\pm$ 0.1 & 64.9 $\pm$ 0.7 & 71.6 $\pm$ 0.3 & ~~~-~~~ & - & \quad~~~-  \\
 \rowcolor{LightYellow}
\msnp (\retune) & 51.9 $\pm$ 0.3 & 61.1 $\pm$ 0.4 & 67.7 $\pm$ 0.3 & 71.7 & 75.7 & \quad80.3 \\
\midrule
\rowcolor{LightCyan}
\ens[16]{\dinop}{\ew} & 55.0 $\pm$ 0.4 \textcolor{Up}{$\mathbf{\uparrow 5.5}$} & 63.4 $\pm$ 0.6 & 69.5 $\pm$ 0.3 & \textbf{73.4} \textcolor{Up}{$\mathbf{\uparrow 2.7}$} & 78.6 & \quad\textbf{81.0} \textcolor{Up}{$\mathbf{\uparrow 0.8}$} \\
\rowcolor{LightYellow}
 \ens[8]{\msnp}{\ew} & \textbf{55.6 $\pm$ 0.2} \textcolor{Up}{$\mathbf{\uparrow 3.7}$} & \textbf{64.5 $\pm$ 0.5} & \textbf{70.3 $\pm$ 0.2} & \textbf{73.4} \textcolor{Up}{$\mathbf{\uparrow 1.7}$}  & \textbf{78.9} & \quad\textbf{80.8} \textcolor{Up}{$\mathbf{\uparrow 0.5}$} \\
\bottomrule
\end{tabular}
}
\end{center}
\vspace{-\baselineskip}
\end{table}

\paragraph{Comparison of different projection heads}
Our method linearly increases projection head parameters, thus a natural question is: Is the gain of $(h_\psi,\mu)$-ensembles due to the increased power (or number of parameters) in projection heads?
We answer this question with an empirical study of non-ensembled projection heads.
Specifically, we studied non-ensembled $h_\psi$ with (depth, width) searched over $\set{2, 3, 4} \times \set{512, 1024, 2048, 4096}$ and measured the linear evaluation performance with different amounts of labeled data. 
In \cref{fig:compare_head_main}, we plot the 1\%-data evaluation result with respect to the number of parameters of the projection head both for ensembled and non-ensembled baselines.
See \cref{appx:study_proj_head} for detailed analysis and extra results for other metrics.
Our key findings are:
\begin{itemize}[leftmargin=15pt,itemsep=0pt,topsep=0pt]
    \item A too powerful non-ensembled $h_\psi$ significantly hurts the label efficiency of learned representations. 
    This result is similar to \citet{chen2020simclrv2}, which found that probing from intermediate layers of projection heads (which can be viewed as using a shallower head) could improve semi-supervised learning (1\%-/10\% labeled data) results. 
    \item The default head (3/2048, denoted as ‘Default’) used in recent SSL methods (SimCLRv2, DINO, MSN, \etc) does not perform as well in few-shot evaluations, probably because it is selected by looking at full-data evaluation metrics.
    In contrast, our baseline (3/1024, denoted as `Our baseline') significantly improves few-shot evaluation performance.
    \item Our $(h_\psi,\mu)$-ensembles outperform all non-ensembled baselines and lead to consistent improvements in all evaluation metrics, despite the increase of parameters.
\end{itemize}





\subsection{Improving SOTA results with ensembleing}
\label{sec:sota}

Next we apply $(h_\psi,\mu)$-ensembles to \dinop and \msnp and compare with the state-of-the-art results. 
We experimented with model architectures \vits/16, \vitb/16, \vitb/8 trained for 800, 400, 300 epochs respectively following \citet{caron2021dino}.
We include both the published results and our retuned versions to ensure strong baselines.
For clarity, we denote our method as ``\{baseline\}-\{ensemble strategy\} (\# of heads)'', \eg, \ens[4]{\dinop}{\ew}.
We tuned both baselines and our methods for all architectures.
We report the best hyperparameters for all models in \cref{appx:best_hyper}.

\cref{table:compare_ens_merge} compares the results of $(h_\psi,\mu)$-ensembles and baselines. 
We find that $(h_\psi,\mu)$-ensembles with \ew consistently improve \emph{all evaluation metrics} (full-data, few-shot) across \emph{both SSL methods} (\dinop, \msnp) and \emph{all architectures} (\vits/16, \vitb/16, \vitb/8) over their non-ensembled counterparts.
The gains in few-shot evaluation are particularly substantial, providing a new state-of-the-art for ImageNet-1K evaluation from ImageNet pretraining.

\subsection{More evaluations for $(h_\psi,\mu)$-ensembles}
\label{sec:transfer}
\begin{table}[h]
\vspace{-.5\baselineskip}
\caption{
    \textbf{Comparison of transfer performance.} 
    \vits/16 is used.
    Our ensemble heads lead to consistent improvements for \msnp and comparable results for \dinop.
}
\label{table:compare_ens_transfer_s16}
\vspace{-\baselineskip}
\begin{center}
\small
\adjustbox{max width=\textwidth}{
\begin{tabular}{lcccccccccc}
\toprule
  & Food101 & CIFAR10 & CIFAR100 & SUN397 & Cars & DTD & Pets & Caltech-101 & Flowers & \rebedit{Avg.}\\
 \midrule
 \dinop & 78.4 & 93.8 & 81.0 & 66.1 & 66.7 & 74.6 & 92.0 & \textbf{94.9} & \textbf{94.4} & \rebedit{82.43} \\
 \ens[16]{\dinop}{\ew} & \textbf{79.1} & 93.8 & \textbf{81.4} & \textbf{66.5} & 66.8 & \textbf{74.9} & \textbf{92.8} & 94.6 & 93.9 & \rebedit{82.64} \\
 \midrule
 \msnp & 77.7 & 93.1 & 79.8 & 64.6 & 63.3 & 72.2 & 92.4 & 94.7 & 92.7 & \rebedit{81.17} \\
 \ens[8]{\msnp}{\ew} & \textbf{78.4} & \textbf{93.9} & \textbf{81.1} & \textbf{65.2} & \textbf{68.0} & \textbf{73.2} & \textbf{93.1} & \textbf{95.4} & \textbf{92.8} & \rebedit{82.34} \\
\bottomrule
\end{tabular}
}
\end{center}
\vspace{-1.5\baselineskip}
\end{table}

\paragraph{Transfer learning}
In \cref{table:compare_ens_transfer_s16}, we compare the transfer learning performance of $(h_\psi,\mu)$-ensembles and non-ensembled baselines.
We used \vits-16 models trained on ImageNet-1K for 800 epochs and evaluated on 9 natural downstream datasets from \citet{chen2020simclr} with linear evaluation (details in \cref{appx:detail_transfer}).
$(h_\psi,\mu)$-ensembles lead to consistent improvements in transfer performance for \msnp and comparable results for \dinop.

\vspace{-.5\baselineskip}
\begin{wraptable}{r}{.35\textwidth}
\vspace{-1.5\baselineskip}
\caption{
    \textbf{Training overhead.}
    Wall-clock time and peak memory per core for training with different numbers of ensembles.
}
\vspace{-\baselineskip}
\label{table:benchmark_comp_cost}
\begin{center}
\small
\adjustbox{max width=\textwidth}{
\begin{tabular}{lcc}
\toprule
 \numens & Wall Time & Peak Memory\\
 \midrule
 1 &  5.81h & 5.25G\\
 4 &  5.91h & 5.40G\\
 16 & 6.34h & 5.89G\\
\bottomrule
\end{tabular}
}
\end{center}
\vspace{-2\baselineskip}
\end{wraptable}
\paragraph{Training overhead}
In \cref{table:benchmark_comp_cost}, we benchmark the computational overhead of $(h_\psi,\mu)$-ensembles at training time.
We used a medium sized model, \dinop with \vitb/16, trained with the same setting used in all of our experiments.
We benchmarked the wall-clock time and peak memory on 128 TPUv3 cores.
$(h_\psi,\mu)$-ensembling is relatively cheap in terms of training cost because the ensembled parts typically account for a small portion of total computation, especially when the backbone encoder is more computationally expensive (\eg, \vitb/8).
Again, we emphasize that there is no evaluation overhead when $(h_\psi,\mu)$-ensembles are removed.


\section{Conclusion \& Discussion}

We introduced an efficient ensemble method for SSL where multiple projection heads are ensembled to effectively improve representation learning.
We showed that with carefully designed ensemble losses that induce diversity over ensemble heads, our method significantly improves recent state-of-the-art SSL methods in various evaluation metrics, particularly for few-shot evaluation. 
\rebedit{Although ensembling is a well-known technique for improving evaluation performance of a single model, we demonstrated that, for models with throw-away parts such as the projection heads in SSL, ensembling these parts can improve the learning of the non-ensembled representation encoder and also achieve significant gains in downstream evaluation without introducing extra evaluation cost.} 

\rebedit{Our ensemble method is applicable to many SSL methods beyond the two we explored. For example, one may consider generalization to BYOL \cite{grill2020byol} or SimSiam \citep{chen2021simsiam} that ensembles projection and/or prediction heads, or MAE \citep{he2022mae} that ensembles the decoders (which introduces more training cost though). 
Our weighted ensemble losses can also be applied as long as the original loss can be reformulated as MLE for some $t$, $s$, and $Y$, \eg, the MSE loss in these methods is MLE under multivariate normal distributions.
We hope our results and insights will motivate more future work for extending our method or exploring more ensemble techniques for SSL.}

In future work, we also hope to remove three limitations of our setting.
First, considering ensembling strategies that include the representation encoder, $r_\omega$, may lead to further improvements in the performance of weighted ensemble SSL, at the cost of increased computation requirements during both training and evaluation.
Second, considering heterogenous architectures in the ensemble may further improve the learned representations (e.g., mixing Transformers with ResNets), whether the heterogeneity is in $r_\omega$, $h_\psi$, or both.
Third, considering other possibilities for $f_{ijy}$ may also reveal performance gains and improve our understanding of the critical aspects that lead to good SSL representations, similar to what we learned about the importance of ensemble diversity.



\subsubsection*{Acknowledgments}
We would like to thank Mathilde Caron and Mahmoud Assran for their extensive help in reproducing DINO and MSN baselines.
We would also like to thank Ting Chen and Yann Dubois for their helpful discussions and encouragements.

\subsubsection*{Reproducibitlity Statement}
We include detailed derivations for all our proposed losses in \cref{app:analysis}.
We report experimental details in \cref{appx:exp_details}, including the implementation details for reproducing the baselines (\cref{appx:repro_improve_baseline}), training and evaluating our methods (\cref{appx:impl_details}), and all hyper-parameters (\cref{appx:best_hyper}) used in our experiments for reproducing our results in \cref{table:compare_ens_merge}.

\bibliography{iclr2023_conference}
\bibliographystyle{iclr2023_conference}

\newpage
\appendix
\section{Pseudocode}
\label{appx:pseudocode}

\begin{algorithm}[h]
\scriptsize
\SetAlgoLined
    \PyComment{b, n, c,: batch size, number of ensemble heads, codebook size} \\
    \PyComment{log\_ps, log\_pt: student, teacher log probabilities with n ensembles}\\
    \PyComment{strategy: ensemble loss average strategy} \\
    \PyComment{tau\_ent: temperature for entropy weighting}\\
    ~\\
    \PyCode{def ensemble\_loss(log\_ps, log\_pt, strategy, tau\_ent):}\\
    \Indp
        \PyCode{b, n, c = log\_pt.shape} \PyComment{axis 1 corresponds to ensemble} \\
        \PyCode{log\_pt = stop\_grad(log\_pt)} \PyComment{stop gradient for teacher}\\
        ~\\
        \PyCode{if strategy == "Unif":}\\
        \Indp
            \PyCode{loss = - (exp(log\_pt) * log\_ps).sum(axis=-1)} \\
            \PyCode{loss = loss.mean(axis=1)} \PyComment{average over ensembles}\\
        \Indm
        \PyCode{elif strategy == "Prob":}\\
        \Indp
            \PyCode{log\_mean\_pt = logsumexp(log\_pt, axis=1, b=1/n)} \PyComment{mean teacher}\\
            \PyCode{log\_mean\_ps = logsumexp(log\_ps, axis=1, b=1/n)} \PyComment{mean student}\\
            \PyCode{loss = - (exp(log\_mean\_pt) * log\_mean\_ps).sum(axis=-1)} \\
        \Indm
        \PyCode{elif strategy == "Ent":}\\
        \Indp
            \PyCode{ent = - (exp(log\_pt) * log\_pt).sum(axis=-1)} \PyComment{teacher entropy} \\
            \PyCode{weight = softmax(-ent/tau\_ent, axis=1)} \PyComment{entropy weights}\\
            \PyCode{loss = - (exp(log\_pt) * log\_ps).sum(axis=-1)} \\
            \PyCode{loss = (loss * weight).sum(axis=1)} \PyComment{entropy weighted average}\\
        \Indm
        ~\\
        \PyCode{return loss.mean()} \PyComment{average over samples} \\
    \Indm
\caption{Pseudocode for computing ensemble loss}
\label{alg:ensemble_loss}
\end{algorithm}

\label{appx:pseudo_code}
\begin{algorithm}[h]
\scriptsize
\SetAlgoLined
    \PyComment{n, c, eta: number of ensemble heads, codebook size, momentum update rate} \\
    \PyComment{fs, ft: student, teacher encoders} \\
    \PyComment{hs\_ens, ht\_ens: student, teacher projection heads with n ensembles, list with length n} \\
    \PyComment{mus\_ens, mut\_ens: student, teacher codebooks with n ensembles, list with length n} \\
    \PyComment{taus, taut: student, teacher temperatures} \\
    \PyComment{strategy: ensemble loss average strategy} \\
    \PyComment{tau\_ent: temperature for entropy weighting} \\
    ~\\
    \PyCode{for x in dataloader:} \PyComment{load a batch with b samples} \\
    \Indp
        \PyCode{xs, xt = augs(x), augt(x)} \PyComment{random augmentations} \\ 
        \PyCode{zs, zt = fs(xs), ft(xt)} 
        \PyComment{representations, (b, l)}\\
        ~\\
        \PyComment{all following computation can be parallelized with batch computation}\\
        \PyCode{log\_ps, log\_pt = [], []}\\
        \PyCode{for j in range(n):}\\
        \Indp
            \PyCode{hs\_j, ht\_j = hs\_ens[j], ht\_ens[j]} \PyComment{j-th projection head}\\
            \PyCode{mus\_j, mut\_j = mus\_ens[j], mut\_ens[j]} \PyComment{j-th codebook, (d, c)}\\
            ~\\
            \PyCode{es\_j, et\_j = hs\_j(zs), ht\_j(zt)} \PyComment{j-th embedding, (b, d)} \\
            ~\\
            \PyCode{rs\_j = (es\_j @ mus\_j) / (es\_j.norm(axis=1, keepdims=True) * mus\_j.norm(axis=0, keepdims=True)) / taus} \PyComment{student logits, (b, c)} \\
            \PyCode{rt\_j = (et\_j @ mut\_j) / (et\_j.norm(axis=1, keepdims=True) * mut\_j.norm(axis=0, keepdims=True)) / taut} \PyComment{teacher logits, (b, c)}\\
            ~\\
            \PyCode{log\_ps\_j = logsoftmax(rs\_j, axis=-1)} \PyComment{(b, c)}\\
            \PyCode{log\_pt\_j = logsoftmax(rt\_j, axis=-1)} \PyComment{(b, c)} \\
            \PyCode{log\_pt\_j = renorm(log\_pt\_j)}
            \PyComment{adjust teacher predictions with centering or sinkhorn, omitted here for simplicity} \\
            ~\\
            \PyCode{log\_ps.append(log\_ps\_j)} \\
            \PyCode{log\_pt.append(log\_pt\_j)} \\
        \Indm
        ~\\
        \PyCode{log\_ps = stack(log\_ps\_j, axis=1)} \PyComment{stacked student log probablities, (b, n, c)}\\
        \PyCode{log\_pt = stack(log\_pt\_j, axis=1)} \PyComment{stacked teacher log probablities, (b, n, c)}\\
        ~\\
        \PyCode{loss = ensemble\_loss(log\_ps, log\_pt, strategy=strategy)} \PyComment{compute ensemble loss} \\
        ~\\
        \PyCode{loss.backward()} \PyComment{back-propagate}\\
        \PyCode{sgd\_update(fs, hs\_ens, mus\_ens)} \PyComment{apply gradient decent update for student}\\
        \PyCode{ema\_update(ft, ht\_ens, mut\_ens, rate=eta)} \PyComment{apply momentum update for teacher}\\
    \Indm
\caption{Pseudocode for ensemble heads with simplified DINO}
\label{alg:ensemble_head_all}
\end{algorithm}

\section{Experimental Details}
\label{appx:exp_details}

In this section, we provide details for our experiments.
In \cref{appx:repro_improve_baseline}, we describe how we reproduced and improved the baseline DINO/MSN models.
We give the implementation details for SSL training and evaluation in \cref{appx:train_details} and \cref{appx:eval_details} respectively.
All the hyper-parameters used in our experiments are in \cref{appx:best_hyper}.

\subsection{Reproducing \& Improving Baselines}
\label{appx:repro_improve_baseline}

We carefully reproduced and further improved baseline methods (denoted as \dinop and \msnp respectively) with an extensive study and hyperparameter search (see \cref{appx:repro_improve_baseline}). 
In particular, we systematically study the projection head design (which we found is crucial for few-shot evaluation performance (\cref{appx:study_proj_head})) and different techniques for avoiding collapse used in both methods (\cref{appx:study_diff_tech_collapse}). 
\dinop performs significantly better than DINO on few-shot evaluation (\eg, 2$\sim$6 percentage point (\pp) gains for 1 shot) and maintains the full-data evaluation performance.
The main adjustments of \dinop are: 
\begin{inlinelist}
\item A 3-layer projection head with a hidden dimension of 1024 (instead of 2048); 
\item Sinkhorn–Knopp (SK) normalization (instead of centering) is applied to teacher predictions, combined with a smaller teacher temperature $\ttemp=0.025$ and codebook size $\numcode=$1024 or 4096. 
\end{inlinelist}
\msnp uses the same projection head as \dinop and applies ME-MAX regularization without SK normalization (which is applied in MSN by default).
Further details for DINO and MSN can be found below.

\subsubsection{DINO}

\begin{table}[h]
\caption{\textbf{Reproducing \& Improving DINO.} Our reproduce results match the public numbers. We further improve the DINO baseline (\dinop) by studying projection heads and collapse-avoiding techniques.
The evaluation results of DINO/\dinop \vits/16 trained with 800 epochs are reported. 
}
\label{table:repro_improve_dino}
\begin{center}
\small
\adjustbox{max width=\textwidth}{
\begin{tabular}{lcccccc}
\toprule
 & \multicolumn{4}{c}{\textbf{Few-shot}} & \multicolumn{2}{c}{\textbf{Full-data}} \\
 \cmidrule(lr){2-5}\cmidrule(lr){6-7}
 & 1 & 2 & 5 & $\sim$13 (1\%) & $k$-NN & Linear \\ 
 \midrule
DINO \citep{caron2021dino} &  38.9 $\pm$ 0.4 & 48.9 $\pm$ 0.3 & 58.5 $\pm$ 0.1 & 64.5 & 74.5 & 76.1 / 77.0 \\
DINO (Ours reproduced) &   39.1 $\pm$ 0.3  &  49.1 $\pm$ 0.5  &  58.6 $\pm$ 0.2  & 64.7 & 74.3 & 75.8 / 76.9 \\
\dinop (\retune) & 44.6 $\pm$ 0.2 & 53.6 $\pm$ 0.3   &  61.1 $\pm$ 0.2  & 66.2 & 74.1 & 75.8 / 76.9 \\
\bottomrule
\end{tabular}
}
\end{center}
\vspace{-\baselineskip}
\end{table}

\paragraph{Reproducing DINO}
\label{appx:validate_impl}
We carefully reproduced DINO with JAX following the official DINO implementation\footnote{\url{https://github.com/facebookresearch/dino}}.
In \cref{table:repro_improve_dino}, we report the evaluation results of DINO using \vits trained with 800 epochs following the exact training configuration for \vits/16 in the official DINO code.
The official results of full-data evaluation and 1\%-data evaluation are from \citet{caron2021dino}, the other few-shot evaluation results are evaluated by \citet{assran2022msn} and also validated by us. 
Note that for consistency of full-data linear evaluation, we report the results with both the \texttt{[CLS]} token representations of the last layer and the concatenation of the \texttt{[CLS]} token representations from the last 4 layers following \citet[][]{caron2021dino}.
For 1-/2-/5-shots evaluation results, we report the mean accuracy and standard deviation across 3 random splits of the data following \citet{assran2022msn}. 
As shown in \cref{table:repro_improve_dino}, our reproduced results are all comparable with the published numbers which validates the implementation of our training and evaluation pipelines. 

\paragraph{Improving DINO}
We improved the DINO baseline with a systematic empirical study of some important components. 
We first empirically compared different techniques for avoiding collapse (see \cref{appx:study_diff_tech_collapse}) and find that Sinkhorn-Knopp (SK) normalization is a more effective and also simpler technique for encouraging codebook usage than the centering operation used in DINO. 
We thus applied SK normalization, which enabled us to use a smaller teacher temperature $\ttemp=0.025$ (instead of $\ttemp=0.07$) and a much smaller codebook size $\numcode=$1024 or 4096 (instead of 65536). These modifications lead to similar performance as DINO with a much smaller codebook (up to 1M parameters, compared to 16M parameters for DINO).
Next we empirically studied the effect of projection heads for different evaluation metrics (see \cref{appx:study_proj_head}), and found that the design of projection heads is crucial for few-shot evaluation metrics and an too power powerful projection head (\eg, the 3-layer MLP with a hidden dimension of 2048 used in DINO/MSN/\etc) could significantly hurt the few-shot performance. With an empirically study of projection head architectures, we found that a simply reducing the hidden dimension to 1024 could significantly improves the few-shot evaluation performance while maintaining full-data evaluation performance. The improved results of \dinop are shown in \cref{table:repro_improve_dino}.

\subsubsection{MSN}
\begin{table}[h]
\caption{\textbf{Reproducing \& improving MSN.} We implement \msnp by adding ME-MAX regularization and masking to \dinop, which surpasses public MSN results.
The evaluation results of MSN/\msnp \vits/16 trained with 800 epochs are reported. }
\label{table:repro_improve_msn}
\begin{center}
\small
\adjustbox{max width=\textwidth}{
\begin{tabular}{lcccccc}
\toprule
 & \multicolumn{4}{c}{\textbf{Few-shot}} & \multicolumn{2}{c}{\textbf{Full-data}} \\
 \cmidrule(lr){2-5}\cmidrule(lr){6-7}
 & 1 & 2 & 5 & $\sim$13 (1\%) & $k$-NN & Linear \\ 
 \midrule
MSN \citep{assran2022msn} &  47.1 $\pm$ 0.1 & 55.8 $\pm$ 0.6 & 62.8 $\pm$ 0.3 & 67.2 & - & ~~~-~~ / 76.9 \\
MSN (\repro) &  39.1 $\pm$ 0.3   &  49.2 $\pm$ 0.3  &  58.4 $\pm$ 0.1  & 64.3 & 72.8 & 74.7 / 75.5 \\
\msnp (\retune) &  47.4 $\pm$ 0.1  & 56.3 $\pm$ 0.4  &  62.8 $\pm$ 0.2   &  67.1  & 73.3 &  75.6 / 76.6 \\ 
\bottomrule
\end{tabular}
}
\end{center}
\vspace{-\baselineskip}
\end{table}
We carefully implemented MSN by adding its main components, \ie, ME-MAX regularization and masking, to the DINO implementation (denoted as \msnp), which surpassed public results as shown in \cref{table:repro_improve_msn}. 
Note that the implementation of \msnp does not exactly match the public implementation in the public MSN code\footnote{\url{https://github.com/facebookresearch/msn}}, where the main differences are: 
\begin{itemize}
    \item MSN applies ME-MAX with Sinkhorn-Knopp normalization by default (as in the released training configuration), which we empirically find does not work very well  (see \cref{table:study_diff_tech_collapse}). \msnp does not apply SK normalization and tunes the regularization strength for ME-MAX.
    \item Some differences in implementation details, \eg, schedules for learning rate/weight decay, batch normalization in projection heads, specific data augmentations, \etc. \msnp uses the exact same setup as \dinop which follows original DINO implementation.
\end{itemize}
We initially tried to exactly reproduce the original MSN following the public MSN code, but the results are much below the public ones, as shown in \cref{table:repro_improve_msn}.
Incorporating the two differences above bridges the gap and makes \msnp surpass the public results.

\subsection{Pretraining details}
\label{appx:train_details}
In this subsection, we provide the general implementation details in \cref{appx:impl_details} and specific hyper-parameters in \cref{appx:best_hyper} in \cref{appx:best_hyper} for reproducibility.

\subsubsection{Implementation Details}
\label{appx:impl_details}

\paragraph{Common setup}
We experimented with DINO \citep{caron2021dino} and MSN \citep{assran2022msn} models on ImageNet ILSVRC-2012 dataset \citep{deng2009imagenet}.
We mainly followed the training setup in \citet{caron2021dino}. In particular, all models were trained with AdamW optimizer \citep{loshchilov2018adamw} and a batch size of 1024. 
The learning rate was linearly warmuped to 0.002 ($=$0.001$\times$batch size/512) and followed a cosine decay schedule. 
The weight decay followed a cosine schedule from 0.04 to 0.4.
The momentum rate for the teacher was increased from 0.996 to 1 with a cosine schedule following BYOL \citep{grill2020byol}.
A stochastic depth \citep{huang2016deep} of 0.1 was applied without dropout \citep{srivastava2014dropout}.
The student temperature $\stemp$ is set to 0.1.
As with DINO, we used the data augmentations of BYOL and multi-crop augmentation of SWAV \citep{caron2020swav}. In particular, 2 global views with a 224$\times$224 resolution and crop area range [0.25, 1.0] were generated for the teacher and student, and another 10 local views with 96$\times$96 resolution and crop area range [0.08, 0.25] were used as extra augmented inputs for the student.
For MSN, we used the exact same setup and incorporated its major component: 1) mean entropy maximization (ME-MAX) regularization; 2) masking as an extra augmentation applied to the student global view.

\paragraph{Main modifications}
We retuned the baselines (\dinop and \msnp) as detailed in \cref{appx:repro_improve_baseline}, and the main adjustments are as followed.
We used a 3-layer projection head with a hidden dimension of 1024.
The output embedding (\ie, $(h_\psi\circ r_\omega)(x)$) and the codes (\ie, $\mu$) both have a dimension of 256 and are $L_2$ normalized.
For \dinop, Sinkhorn–Knopp (SK) normalization was applied to teacher predictions.
For \msnp, ME-MAX was used without SK normalization
and the regularization strength was tuned over \{3, 4, 5\}.
For all models, we used teacher temperature $\ttemp=0.025$ which was linearly decayed from 0.05 for the first 30 epochs. 
The codebook size $\numcode$ was selected over \{1024, 4096\} for all models, and typically $\numcode=$4096 was selected for baseline methods and $\numcode=$1024 was selected for ours.
For our $(h_\psi,\mu)$-ensembles with \ew, entropy weighting temperature $\enttemp$ is linearly decayed from 0.5 to the specified value.

\subsubsection{Hyper-parameters}
\label{appx:best_hyper}
We report the hyperparameters for training our models for reproducibility:

\begin{table}[h]
\caption{Hyper-parameters for training the \dinop model.}
\label{table:hyperparam_dino}
\begin{center}
\scriptsize
\adjustbox{max width=\textwidth}{
\begin{tabular}{lccccccc}
\toprule
 \multirow{2}{*}{Hyper-parameter} & \multicolumn{3}{c}{\vits/16} & \multicolumn{2}{c}{\vitb/16} & \multicolumn{2}{c}{\vitb/8} \\
 \cmidrule(lr){2-4}
 \cmidrule(lr){5-6}\cmidrule(lr){7-8}
& \dinop & \ens[16]{\dinop}{\inside} & \ens[4/16]{\dinop}{\ew} & \dinop & \ens[16]{\dinop}{\ew} & \dinop & \ens[16]{\dinop}{\ew} \\
 \midrule
training epoch & \multicolumn{3}{c}{800} & \multicolumn{2}{c}{400} & \multicolumn{2}{c}{300}\\
batch size & \multicolumn{3}{c}{1024} & \multicolumn{2}{c}{1024} & \multicolumn{2}{c}{1024} \\
learning rate & \multicolumn{3}{c}{2e-3} & \multicolumn{2}{c}{2e-3} & \multicolumn{2}{c}{2e-3} \\
warmup epoch & \multicolumn{3}{c}{10} & \multicolumn{2}{c}{30} & \multicolumn{2}{c}{10}\\
min lr & \multicolumn{3}{c}{1e-5} & \multicolumn{2}{c}{1e-5} & \multicolumn{2}{c}{4e-5} \\
weight decay & \multicolumn{3}{c}{0.04 $\to$ 0.4} & \multicolumn{2}{c}{0.04 $\to$ 0.4} & \multicolumn{2}{c}{0.04 $\to$ 0.4} \\
stochastic depth & \multicolumn{3}{c}{0.1} & \multicolumn{2}{c}{0.1} & \multicolumn{2}{c}{0.1}\\
gradient clip & \multicolumn{3}{c}{3.0} & \multicolumn{2}{c}{1.0} & \multicolumn{2}{c}{3.0} \\
\midrule
momentum & \multicolumn{3}{c}{0.996 $\to$ 1.0} & \multicolumn{2}{c}{0.996 $\to$ 1.0} & \multicolumn{2}{c}{0.996 $\to$ 1.0}\\
\# of multi-crops & \multicolumn{3}{c}{10} & \multicolumn{2}{c}{10} & \multicolumn{2}{c}{10}\\
masking ratio & \multicolumn{3}{c}{-} & \multicolumn{2}{c}{-} & \multicolumn{2}{c}{-} \\
\midrule
proj. layer & \multicolumn{3}{c}{3} & \multicolumn{2}{c}{3} & \multicolumn{2}{c}{3}\\
proj. hidden dim & \multicolumn{3}{c}{1024} & \multicolumn{2}{c}{1024} & \multicolumn{2}{c}{1024} \\
emb. dim $\embdim$ & \multicolumn{3}{c}{256} & \multicolumn{2}{c}{256} & \multicolumn{2}{c}{256} \\
rep. dim & \multicolumn{3}{c}{384} & \multicolumn{2}{c}{768} & \multicolumn{2}{c}{768} \\
codebook size $\numcode$ & 4096 & 1024 & 1024 & 4096 & 1024  & 4096 & 1024 \\

\midrule
student temp. & \multicolumn{3}{c}{0.1} & \multicolumn{2}{c}{0.1} & \multicolumn{2}{c}{0.1} \\
teacher temp. & \multicolumn{3}{c}{0.025} & \multicolumn{2}{c}{0.025} & \multicolumn{2}{c}{0.025} \\
te. temp. decay epoch & \multicolumn{3}{c}{30} & \multicolumn{2}{c}{30} & \multicolumn{2}{c}{30} \\
center & \multicolumn{3}{c}{\xmark} & \multicolumn{2}{c}{\xmark} & \multicolumn{2}{c}{\xmark} \\
SK norm & \multicolumn{3}{c}{\cmark} & \multicolumn{2}{c}{\cmark} & \multicolumn{2}{c}{\cmark} \\
ME-MAX weight & \multicolumn{3}{c}{-} & \multicolumn{2}{c}{-} & \multicolumn{2}{c}{-} \\
\midrule
ent. weight temp. $\enttemp$ & - & - & 0.05 & - & 0.05 &  - & 0.06\\
$\enttemp$ init. & - & - & 0.5 & - & 0.5 & - & 0.5 \\
$\enttemp$ decay epoch & - & - & 30 & - & 30 & - & 30 \\

\bottomrule
\end{tabular}
}
\end{center}
\vspace{-\baselineskip}
\end{table}
\begin{table}[h]
\caption{Hyper-parameters for training the \msnp model.}
\label{table:hyperparam_mnp}
\begin{center}
\scriptsize
\adjustbox{max width=\textwidth}{
\begin{tabular}{lcccccc}
\toprule
 \multirow{2}{*}{Hyper-parameter} & \multicolumn{2}{c}{\vits/16} & \multicolumn{2}{c}{\vitb/16} & \multicolumn{2}{c}{\vitb/8} \\
 \cmidrule(lr){2-3}
 \cmidrule(lr){4-5}\cmidrule(lr){6-7}
& \dinop & \ens[2/8]{\msnp}{\ew} & \msnp & \ens[8]{\msnp}{\ew} & \msnp & \ens[8]{\msnp}{\ew} \\
 \midrule
training epoch & \multicolumn{2}{c}{800} & \multicolumn{2}{c}{400} & \multicolumn{2}{c}{300}\\
batch size & \multicolumn{2}{c}{1024} & \multicolumn{2}{c}{1024} & \multicolumn{2}{c}{1024} \\
learning rate & \multicolumn{2}{c}{2e-3} & \multicolumn{2}{c}{2e-3} & \multicolumn{2}{c}{2e-3} \\
warmup epoch & \multicolumn{2}{c}{20} & \multicolumn{2}{c}{30} & \multicolumn{2}{c}{20}\\
min lr & \multicolumn{2}{c}{1e-5} & \multicolumn{2}{c}{4e-5} & \multicolumn{2}{c}{4e-5} \\
weight decay & \multicolumn{2}{c}{0.04 $\to$ 0.4} & \multicolumn{2}{c}{0.04 $\to$ 0.4} & \multicolumn{2}{c}{0.04 $\to$ 0.4} \\
stochastic depth & \multicolumn{2}{c}{0.1} & \multicolumn{2}{c}{0.1} & \multicolumn{2}{c}{0.1}\\
gradient clip & \multicolumn{2}{c}{1.0} & \multicolumn{2}{c}{1.0} & \multicolumn{2}{c}{1.0} \\
\midrule
momentum & \multicolumn{2}{c}{0.996 $\to$ 1.0} & \multicolumn{2}{c}{0.996 $\to$ 1.0} & \multicolumn{2}{c}{0.996 $\to$ 1.0}\\
\# of multi-crops & \multicolumn{2}{c}{10} & \multicolumn{2}{c}{10} & \multicolumn{2}{c}{10}\\
masking ratio & \multicolumn{2}{c}{0.2} & \multicolumn{2}{c}{0.2} & \multicolumn{2}{c}{0.15} \\
\midrule
proj. layer & \multicolumn{2}{c}{3} & \multicolumn{2}{c}{3} & \multicolumn{2}{c}{3}\\
proj. hidden dim & \multicolumn{2}{c}{1024} & \multicolumn{2}{c}{1024} & \multicolumn{2}{c}{1024} \\
emb. dim $\embdim$ & \multicolumn{2}{c}{256} & \multicolumn{2}{c}{256} & \multicolumn{2}{c}{256} \\
rep. dim & \multicolumn{2}{c}{384} & \multicolumn{2}{c}{768} & \multicolumn{2}{c}{768} \\
codebook size $\numcode$ & 4096 & 1024 & 4096 & 1024  & 4096 & 1024 \\

\midrule
student temp. & \multicolumn{2}{c}{0.1} & \multicolumn{2}{c}{0.1} & \multicolumn{2}{c}{0.1} \\
teacher temp. & \multicolumn{2}{c}{0.025} & \multicolumn{2}{c}{0.025} & \multicolumn{2}{c}{0.025} \\
te. temp. decay epoch & \multicolumn{2}{c}{30} & \multicolumn{2}{c}{30} & \multicolumn{2}{c}{30} \\
center & \multicolumn{2}{c}{\xmark} & \multicolumn{2}{c}{\xmark} & \multicolumn{2}{c}{\xmark} \\
SK norm & \multicolumn{2}{c}{\xmark} & \multicolumn{2}{c}{\xmark} & \multicolumn{2}{c}{\xmark} \\
ME-MAX weight & \multicolumn{2}{c}{4.0} & \multicolumn{2}{c}{4.0} & \multicolumn{2}{c}{4.0} \\
\midrule
ent. weight temp. $\enttemp$ & - & 0.01 & - & 0.005 &  - & 0.01 \\
$\enttemp$ init. & - & 0.5 & - & 0.5 & - & 0.5 \\
$\enttemp$ decay epoch & - & 30 & - & 30 & - & 30 \\

\bottomrule
\end{tabular}
}
\end{center}
\vspace{-\baselineskip}
\end{table}

\subsection{Evaluation Protocals}
\label{appx:eval_details}
\paragraph{Few-shot linear evaluation}
We followed the few-shot evaluation protocal in \citet{assran2022msn}. Specifically, we used the 1-/2-/5-shot ImageNet dataset splits\footnote{Publicly available at \url{https://github.com/facebookresearch/msn}} in \citet{assran2022msn} and 1\% ($\sim$13-shot) ImageNet dataset splits\footnote{Publicly available at \url{https://github.com/google-research/simclr/tree/master/imagenet_subsets}}. For given labelled images, we took a single central crop of size $224 \times 224$ without additional data augmentations, and extracted the output \texttt{[CLS]} token representations from the frozen pretrained model. Then we trained a linear classifier with multi-class logistic regression on top of the extracted representations. We used the \texttt{scikit-learn} package \citep{scikit-learn} for the logistric regression classifier. For all few-shot evaluations, we searched the $\text{L}_2$ regularization strength over \{1e-4, 3e-4, 1e-3, 3e-3, 1e-2, 3e-2, 1e-1, 3e-1, 1, 3, 10\}.

\paragraph{Full-data linear evaluation}
We followed the linear evaluation protocal in \cite{caron2021dino}. Specifically, we trained a linear classifier on top of the representations extracted from the frozen pretrained model. The linear classifier is optimized by SGD with Nesterov momentum \citep{Nesterov1983nag,sutskever2013importance} of 0.9 and a batch size of 4096 for 100 epochs on the whole ImageNet dataset, following a cosine learning rate decay schedule.
We did not apply any weight decay.
During training, we only applied basic data augmentations including random resized crops of size $224 \times 224$ and horizontal flips. During testing, we took a single central crop of the same size. For \vits/16, \citet{caron2021dino} found that concatenating the \texttt{[CLS]} token representations from the last $l$ (specifically, $l=4$) layers (\cf Appendix F.2 in \citet{caron2021dino}) improved the results by about 1 \pp. We followed the same procedure, but reported linear evaluation results with both $l=1$ and $l=4$ in \cref{table:compare_ens_merge} for consistency.
In our empirical study with \vits/16, we used the result with $l=1$.
For larger models (\eg, \vitb/16), we followed \citet{caron2021dino, zhou2022ibot} to use the concatenation of the \texttt{[CLS]} token representation and the average-pooled patch tokens from the last $l=1$ layer for linear evaluation.
For all linear evaluations, we searched the base learning rate over \{4.8e-3, 1.6e-2, 4.8e-2, 1.6e-1, 4.8e-1, 1.6\}. 

\paragraph{Full-data $k$-NN evaluation}
We followed the $k$-NN evaluation protocal in \citet{caron2021dino,wu2018instdisc}. Specifically, for each image in the given dataset, we took a single central crop of size $224 \times 224$ without additional data augmentations, and extracted the output \texttt{[CLS]} token representations from the frozen pretrained model. 
The extracted representations are used for a weighted $k$-Nearest-Neighbor classifier. In particular, denote the stored training representations and labels as $\mathcal{D}=\set{(z_i, y_i)}_{i=1}^N$. For a test image with extracted representation $z$, denote the set of its top $k$-NN training samples as $\mathcal{D}_k[z] \subseteq \mathcal{D}$ and $\norm{\mathcal{D}_k[z]} = k$. The $k$-NN set $\mathcal{D}_k[z]$ is used to make the prediction for the test image with a weighted vote, \ie, $\hat{y}=\argmax_y \pa{\sum_{(z_j, y_j) \in \mathcal{D}_k[z]} \alpha_j \mathbf{1}_{y=y_j}}$, where $\mathbf{1}_{y=y_j}$ is the one-hot vector corresponding to label $y_j$ and $\alpha_j$ is the weight induced by the cosine similarity between $z$ and $z_j$, \ie, $\alpha_j = \exp \pa{ \frac{1}{\tau'}\frac{z^\T z_j}{ \normdouble{z} \normdouble{z_j}}}$. We set $\tau' = 0.07$ without tuning as in \citet{caron2021dino,wu2018instdisc}. For all $k$-NN evaluations, we searched $k$ over \{5, 10, 20, 50, 100\} and found that $k=10$ or $k=20$ was consistently the best.

\paragraph{Transfer evaluation via linear probing}
\label{appx:detail_transfer}
We mainly followed the transfer evaluation protocal in \cite{grill2020byol,chen2020simclr}. In particular, we used 9 of their 13 datasets that are available in \texttt{tensorflow-datasets} \citep{tfds}, namely Food-101 \citep{bossard2014food}, CIFAR10 \citep{krizhevsky2009cifar}, CIFAR100 \citep{krizhevsky2009cifar}, SUN397 scene dataset \citep{xiao2010sun}, Stanford Cars \citep{krause2013cars},  Describable Textures Dataset \citep[DTD]{cimpoi2014dtd}, Oxford-IIIT Pets \citep{parkhi2012pets}, Caltech-101 \citep{fei2004caltech101}, Oxford 102 Flowers \citep{nilsback2008flowers}. Following their evaluation metrics, we reported mean per-class accuracy for Oxford-IIIT Pets, Caltech-101, and Oxford 102 Flowers datasets and reported top-1 accuracy for other datasets. 
We transferred the models pretrained on ImageNet \citep{deng2009imagenet} to these datasets by training a linear classifer on top of frozen representations.
In particular, we resized given images to $256 \times 256$  and took a single central crop of size $224 \times 224$ without additional data augmentations. We extracted the output \texttt{[CLS]} token representations from the frozen pretrained model. Then we trained a linear classifier with multi-class logistic regression on top of the extracted representations. We used the \texttt{scikit-learn} package \citep{scikit-learn} for the logistric regression classifier. For all transfer evaluations, we searched the $\text{L}_2$ regularization strength over \{1e-6, 1e-5, 1e-4, 3e-4, 1e-3, 3e-3, 1e-2, 3e-2, 1e-1, 3e-1, 1, 3, 1e1, 3e1, 1e2, 1e3, 1e4, 1e5\}.

\section{Additional Results}
\label{appx:additional_results}

\subsection{Empirical study of techniques for avoiding collapse}
\label{appx:study_diff_tech_collapse}
\begin{table}[t]
\caption{\textbf{Empirical study of different techniques for avoiding collapse.} Using Sinkhorn-Knopp normalization instead of centering for DINO leads to improved performance, and matches the original DINO even with a much smaller codebook. The ME-MAX regularization of MSN is very effective and leads to significant improvement for few-shot evaluations.}
\label{table:study_diff_tech_collapse}
\begin{center}
\small
\adjustbox{max width=\textwidth}{
\begin{tabular}{lccccccccc}
\toprule
 & \multicolumn{3}{c}{\textbf{Technique}} & \multicolumn{4}{c}{\textbf{Few-shot}} & \multicolumn{2}{c}{\textbf{Full-data}} \\
 \cmidrule(lr){2-4}\cmidrule(lr){5-8}\cmidrule(lr){9-10}
 & Center &  Sinkhorn & ME-MAX & 1 & 2 & 5 & $\sim$13 (1\%) & $k$-NN & Linear \\ 
 \midrule
\multirow{2}{*}{DINO} & \checkmark &  & & 37.8 $\pm$ 0.4  & 47.4 $\pm$ 0.3 & 56.9 $\pm$ 0.4 & 63.0 & 72.4 & 74.9 \\
 & & \checkmark &  &  39.1 $\pm$ 0.3 &  49.4 $\pm$ 0.3 & 58.7 $\pm$ 0.2  & 64.8 & 74.1 & 76.0 \\
  \cmidrule(lr){1-10}
\multirow{2}{*}{MSN} & & \checkmark & \checkmark & 36.0 $\pm$ 0.4  & 46.6 $\pm$ 0.6  & 56.5 $\pm$ 0.2  & 63.2 & 73.2 & 75.2 \\
 & &  & \checkmark & 43.9 $\pm$ 0.2  & 53.0 $\pm$ 0.3  & 61.1 $\pm$ 0.2  & 66.0 & 74.0 & 75.8 \\
\bottomrule
\end{tabular}
}
\end{center}
\vspace{-\baselineskip}
\end{table}


\begin{table}[t]
\caption{ME-MAX regularization is sensitive to hyper-parameters.}
\label{table:study_memax}
\begin{center}
\small
\adjustbox{max width=\textwidth}{
\begin{tabular}{cccccccc}
\toprule
 \multirow{2}{*}{\textbf{Weight}} &  \multicolumn{4}{c}{\textbf{Few-shot}} & \multicolumn{2}{c}{\textbf{Full-data}} \\
 \cmidrule(lr){2-5}\cmidrule(lr){6-7}
 & 1 & 2 & 5 & $\sim$13 (1\%) & KNN & Linear \\ 
 \midrule
1.0  & 37.6 $\pm$ 0.2 & 48.0 $\pm$ 0.4 & 57.7 $\pm$ 0.2 & 64.0 & 73.5 & 75.6 \\
3.0 & 43.9 $\pm$ 0.2  & 53.0 $\pm$ 0.3  & 61.1 $\pm$ 0.2  & 66.0 & 74.0 & 75.8 \\
5.0 & 43.6 $\pm$ 0.2 & 52.6 $\pm$ 0.4 & 60.4 $\pm$ 0.1 & 65.5 & 73.9 & 75.6 \\

\bottomrule
\end{tabular}
}
\end{center}
\vspace{-\baselineskip}
\end{table}

Most self-supervised learning methods utilize some techniques to avoid collapse of representations with, \eg, contrastive loss \citep{chen2020simclr,he2020moco}, batch normalization \cite{grill2020byol}, asymmetric architecture design with a predictor \cite{grill2020byol,chen2021simsiam}, \etc. In DINO and MSN, a learnable codebook is used for the learning objective and different techniques are applied to encourage the effective codebook usage. There are two potential cases of collapse (as discussed in \citet{caron2021dino}):
\begin{itemize}
    \item \textit{Dominating codes}. This is the case of ``winner-take-all'': only a small portion of codes are being predicted while others are inactive. Typical solutions for avoiding this include applying Sinkhorn–Knopp normalization \citep{cuturi2013sinkhorn} as in SWaV \citep{caron2020swav}, centering teacher logits as in DINO \citep{caron2021dino}, and applying mean-entropy maximization regularization (ME-MAX) as in MSN \citep{assran2022msn}. Note that in MSN, ME-MAX is combined with Sinkhorn–Knopp normalization by default.
    \item \textit{Uniform codes}. This is the case where all codes are treated equally and the predictions reduce to be uniform over codes. A simple and effective solution is to applying sharpening, \ie, using a lower temperature for computing the teacher prediction. 
\end{itemize}

We systematically study different techniques in a unified setup. In particular, we used DINO with the \vits backbone, a 3-layer MLP projection head with hidden dimension 2048, and a codebook of size 4096 and dimension 256. We applied different techniques to DINO and searched the teacher temperature in $\set{0.0125, 0.025, 0.05}$ for each. 
For ME-MAX, we searched regularization weight in $\set{1.0, 3.0, 5.0}$.
For ME-MAX combined with Sinkhorn, we followed \citet{assran2022msn} and used default regularization weight of 1.0.
The results are in \cref{table:study_memax}. We observed that:
\begin{itemize}
    \item DINO's centering operation is not as strong as other techniques, and it favours a larger teacher temperature (\eg, 0.05). It does not work well when the codebook size (4096) is not as large as the one used in the original DINO model (65536). Switching to use Sinkhorn–Knopp normalization leads to much more improved performance, and matches the performance of original DINO (\cref{table:repro_improve_dino}) with a much smaller codebook.
    \item MSN's ME-MAX regularization is very effective, and leads to significant improvements over others. We also found it is sensitive to the regularization weight and teacher temperature (\cf \cref{table:study_memax}). However, we observed that combining ME-MAX with Sinkhorn does not work well without tuning the regularization weight (which is recommended by \citet{assran2022msn}).
\end{itemize}

\subsection{Empirical Study of Projection Heads}
\label{appx:study_proj_head}

\begin{figure}[t]
    \centering
    \begin{subfigure}[b]{0.45\linewidth}
        \includegraphics[width=\linewidth, trim={0 0 0 0}, clip]{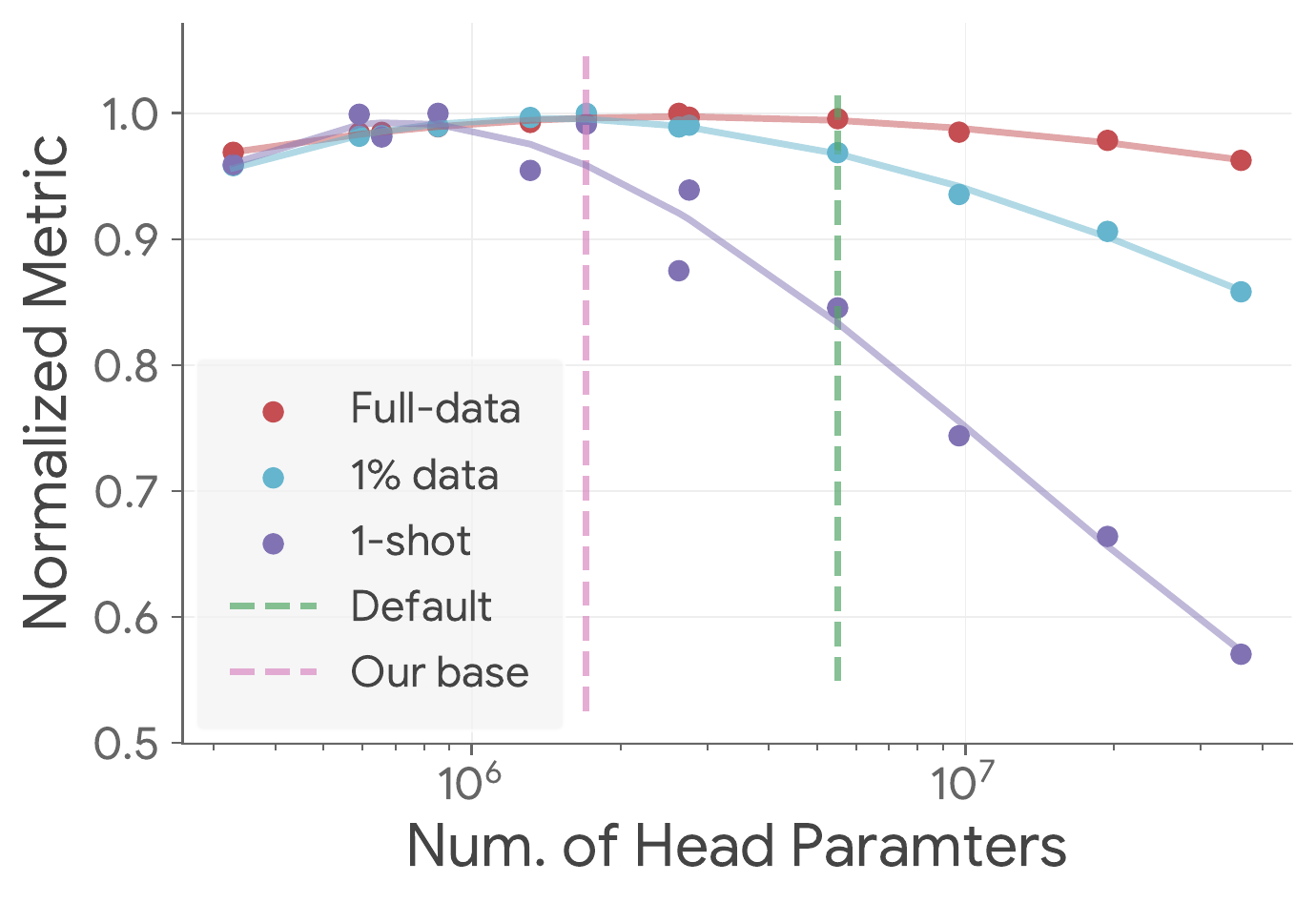}
        \vspace{-\baselineskip}
        \caption{{Merged}}
        \label{fig:compare_head_merge}
    \end{subfigure}
    \begin{subfigure}[b]{0.45\linewidth}
        \includegraphics[width=\linewidth, trim={0 0 0 0}, clip]{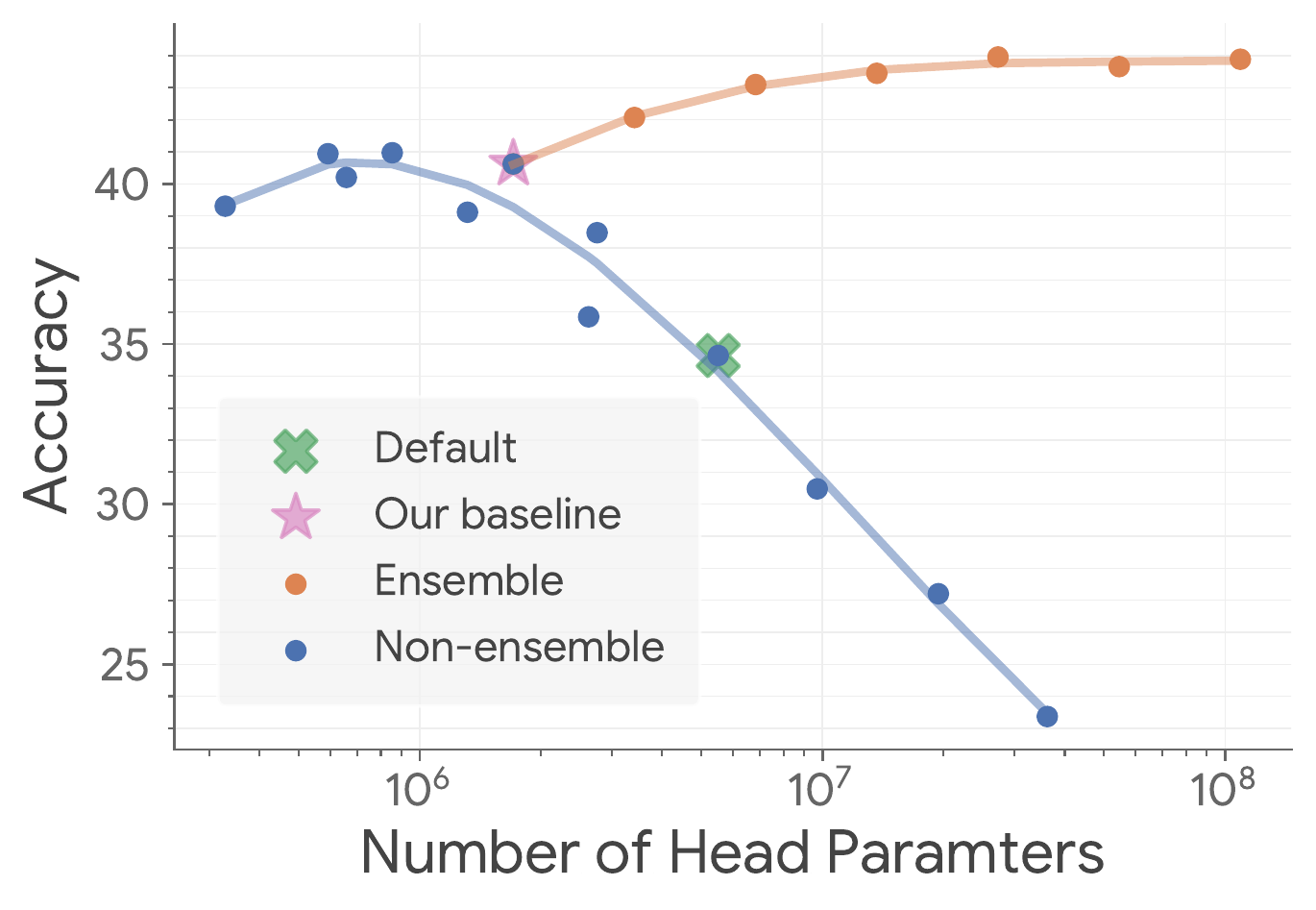}
        \vspace{-\baselineskip}
        \caption{1-shot}
        \label{fig:compare_head_1shot}
    \end{subfigure}
    \hspace{1em}
    \begin{subfigure}[b]{0.45\linewidth}
        \includegraphics[width=\linewidth, trim={0 0 0 0}, clip]{figures/plots/plot_eval_metric_vs_diff_head_1pct.pdf}
        \vspace{-\baselineskip}
        \caption{{1\%-data}}
        \label{fig:compare_head_1pct}
    \end{subfigure}
    \hspace{1em}
    \begin{subfigure}[b]{0.45\linewidth}
        \includegraphics[width=\linewidth, trim={0 0 0 0}, clip]{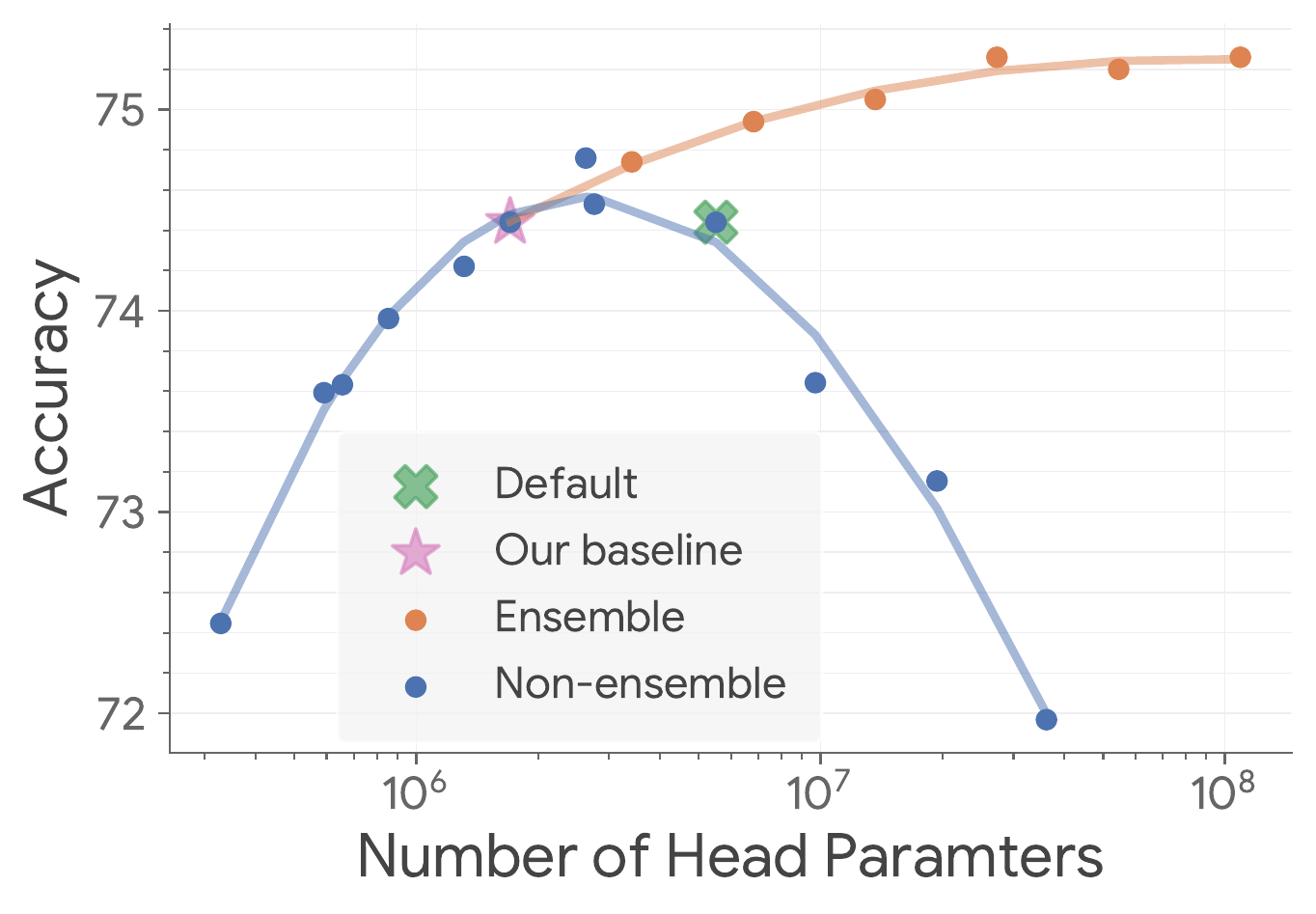}
        \vspace{-\baselineskip}
        \caption{{Full-data}}
        \label{fig:compare_head_full}
    \end{subfigure}
    \caption{\textbf{Effect of projection heads for different evaluation metrics}. We compare non-ensemble projection heads with different depths and widths as well as our $(h_\psi,\mu)$-ensembles, and evaluate linear evaluation performance with different amount of labeled data. (a) shows the comparison of normalized metrics for non-ensembles. (b)-(d) compares non-ensemble and $(h_\psi,\mu)$-ensembles by unnormalized metrics. `Default' denotes the default projection heads used in many SSL methods. See analysis in \cref{appx:study_proj_head} for details.}
    \label{fig:compare_head_all}
\end{figure}

In this subsection, we systematically study the effect of projection heads for different evaluation metrics. In particular, we used \dinop \vits/16 as the base model and used different projection heads with (depth, width) searched over $\set{2, 3, 4}\times \set{512, 1024, 2048, 4096}$. All models are trained with 300 epochs using exact the same set of hyper-parameters.
We measured the linear evaluation performance with different amount of labeled data (\ie, full-data, 1\%-data, 1-shot).

In \cref{fig:compare_head_merge}, we plot different evaluation metrics (normalized respectively by the best of each) versus the number of projection head parameters.
In \cref{fig:compare_head_1shot,fig:compare_head_1pct,fig:compare_head_full}, we plot each unnormalized evaluation metric respectively for different heads as well as our $(h_\psi,\mu)$-ensembles.
Our key findings are:
\begin{itemize}
    \item The projection head has a relatively larger impact on few-shot evaluation metrics, as reflected by the relative magnitudes of different metrics in \cref{fig:compare_head_merge}.
    An too powerful non-ensemble projection head \emph{significantly hurts the label efficiency} of learned representations, reflected by a much larger drop in few-shot evaluation performance (up to 18 \pp for 1-shot, 9 \pp for 1\%-data).
    This result is also partially observed in \citet{chen2020simclrv2}, where they found that probing from intermediate layers of projection heads (which can be viewed as using a shallower head) could improve the semi-supervised learning (1\%-/10\%) results. 
    \item The optimal projection head for different metrics can differ a lot. A weaker head improves label efficiency (few-shot performance), while a stronger (but not too strong) head improves linear decodability.
    As a result, the default projection head (3/2048) that is widely used in SimCLR v2 \citep{chen2020simclrv2}, DINO \cite{caron2021dino}, iBOT \citep{zhou2022ibot}, MSN \citep{assran2022msn}, \etc, does not perform well in few-shot evaluations (as shown by the green cross denoted as `Default'), probably because it is selected by full-data evaluation metrics.
    \item There exist some projection heads that performs decently well on all evaluation metrics, \eg, the baseline model (3/1024) used in our experiments (pink star denoted as `Our base').
    \item Compared to naively tuning projection head architectures, our $(h_\psi,\mu)$-ensembles (orange curves in \cref{fig:compare_head_1shot,fig:compare_head_1pct,fig:compare_head_full}) consistently improve all metrics with different amount of labeled data, despite it also increases the number of parameters in projection heads.
    Our $(h_\psi,\mu)$-ensembles outperform all non-ensembles, which also include the counterparts of probing from intermediate layers from the a deeper head (\ie, shallower heads).
\end{itemize}

\subsection{Empirical study of $(h_\psi,\mu)$-ensembles}
\label{appx:study_ens_head}

\begin{figure}[t]
    \centering
    \includegraphics[width=0.5\textwidth]{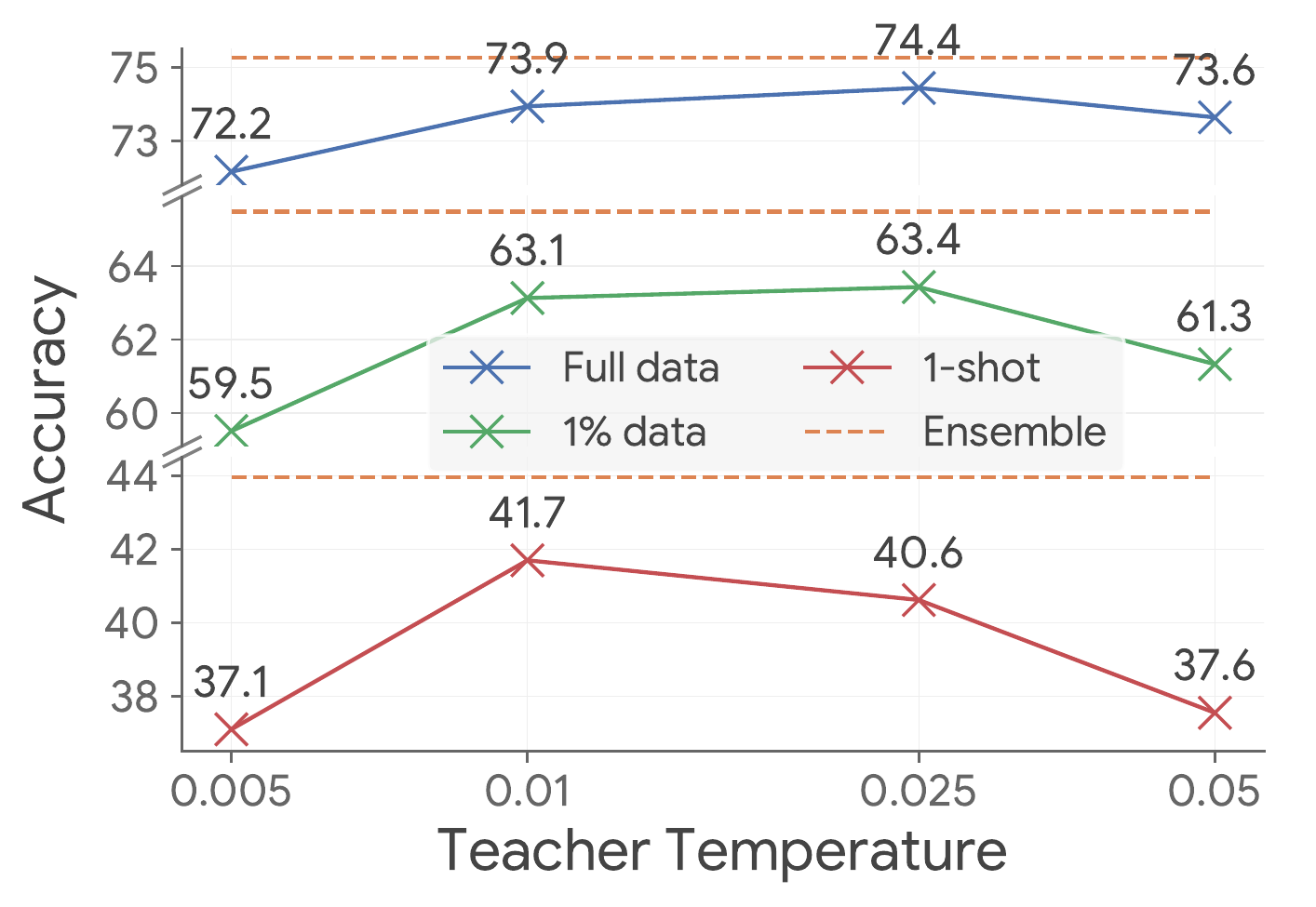}
    \caption{\textbf{Effect of teacher temperature} for non-ensemble \dinop. 
    \dinop with a lower temperature can achieve better few-shot performance, but still under-performs our ensemble method (\ens{\dinop}{\ew} with 16 heads, orange lines). \dinop \vits/16 trained for 300 epochs is used and $\ttemp=0.025$ is used for \ens{\dinop}{\ew}.}
    \label{fig:study_teacher_temp_baseline}
\end{figure}
\paragraph{Are the gains of \ew purely from sharper teacher predictions?}
Our \ew strategy assigns higher weights to the heads that predict with lower entropies, thus effectively uses sharper teacher predictions as the targets.
One may be curious about how this effect accounts for the gains of the \ew strategy.
We empirically answer this question by studying the non-ensemble baseline that uses a sharper teacher predictions in a \emph{data-independent} manner (in contrast to \ew, which uses \emph{data-dependent} entropy weights). 
Specifically, we compare the non-ensemble \dinop that use different teacher temperature $\ttemp \in \set{0.005, 0.01, 0.025, 0.05}$ and also our \ens[16]{\dinop}{\ew} with $\ttemp=0.025$, as shown in \cref{fig:study_teacher_temp_baseline}.
We find that the teacher temperature has a big impact on evaluation results especially for few-shot evaluation. 
Compared to our default baseline that uses $\ttemp=0.025$, a lower temperature (\eg, $\ttemp=0.01$) can indeed improve the 1-shot performance (at the cost of worse full-data performance). However, an too low temperature ($\ttemp=0.005$) will hurt the performance. 
Our \ens[16]{\dinop}{\ew} consistently outperform all the baselines, which implies the importance of selecting sharper teacher predictions in a data-dependent manner.

\paragraph{Comparison of different ensemble strategies and variants}
\begin{table}[t]
\caption{\textbf{Full table of \cref{table:compare_ens_ablation}} including all metrics for comparing different ensemble strategies. \ew and \inside significantly improves over the non-ensemble baseline, while \outside leads to no gains.
Ensembling the whole projection head works the best.
All models are \dinop \vits/16 trained for 300 epochs.
The means and standard deviations over 3 initialization seeds for all evaluation results are reported.
}
\label{table:compare_ens_ablation_full}
\begin{center}
\small
\adjustbox{max width=\textwidth}{
\begin{tabular}{lccccccccc}
\toprule
\multirow{2}{*}{\textbf{How}} & \multicolumn{2}{c}{\textbf{Where}} & \multicolumn{4}{c}{\textbf{Few-shot}} & \multicolumn{2}{c}{\textbf{Full-data}} \\
 \cmidrule(lr){2-3}\cmidrule(lr){4-7}\cmidrule(lr){8-9}
  & Proj. Head &  Codebook & 1 & 2 & 5 & $\sim$13 (1\%) & $k$-NN & Linear \\ 
 \midrule
 Base & & & 40.6 $\pm$ 0.2 & 49.8 $\pm$ 0.2 & 57.9 $\pm$ 0.3 & 63.4 $\pm$ 0.2 & 72.3 $\pm$ 0.1 & 74.4 $\pm$ 0.1\\
 \midrule 
\outside & \checkmark &  \checkmark & 40.4 $\pm$ 0.4 & 49.5 $\pm$ 0.4 & 57.6 $\pm$ 0.3 &  63.3 $\pm$ 0.3 & 72.2 $\pm$ 0.2 & 74.5 $\pm$ 0.2\\
\midrule
\inside & \checkmark &   &  39.7 $\pm$ 0.5 & 49.0 $\pm$ 0.5 & 57.4 $\pm$ 0.4 & 63.0 $\pm$ 0.4 & 72.8 $\pm$ 0.2 & 74.8 $\pm$ 0.1 \\
\inside & \checkmark &  \checkmark & 41.9 $\pm$ 0.3 & 51.5 $\pm$ 0.5 & 59.6 $\pm$ 0.4 & 65.1 $\pm$ 0.3 & \textbf{73.7 $\pm$ 0.3} & \textbf{75.4 $\pm$ 0.1} \\
\midrule
\ew &  &  \checkmark &  40.6 $\pm$ 0.4 & 49.5 $\pm$ 0.6 & 58.0 $\pm$ 0.4 & 63.5 $\pm$ 0.4 & 72.1 $\pm$ 0.3 & 74.5 $\pm$ 0.3 \\
\ew & \checkmark &   & 43.0 $\pm$ 0.6 & 52.2 $\pm$ 0.8 & 59.7 $\pm$ 0.7 & 64.8 $\pm$ 0.5 & 72.9 $\pm$ 0.6 & 75.1 $\pm$ 0.4  \\
\ew & \checkmark &  \checkmark & \textbf{44.0 $\pm$ 0.2} & \textbf{53.0 $\pm$ 0.5} & \textbf{60.5 $\pm$ 0.3} & \textbf{65.5 $\pm$ 0.1} & 73.2 $\pm$ 0.1 & \textbf{75.3 $\pm$ 0.1}  \\
\midrule 
\ew-\textsc{St} & \checkmark &  \checkmark & 40.0 $\pm$ 0.5 & 39.2 $\pm$ 0.6 & 57.3 $\pm$ 0.5 & 62.7 $\pm$ 0.5 & 71.9 $\pm$ 0.4 & 74.0 $\pm$ 0.4  \\
\bottomrule
\end{tabular}
}
\end{center}
\vspace{-\baselineskip}
\end{table}
\begin{table}[t]
\caption{\textbf{Comparison of different varaints of \inside.} The \inside strategy used in our experiments performs the best. '-' in the table denotes training divergence for \inside-\textsc{Max}. The experimental setup is the same as \cref{table:compare_ens_ablation_full}.
}
\label{table:compare_ens_prob_variant}
\begin{center}
\small
\adjustbox{max width=\textwidth}{
\begin{tabular}{lccccccccc}
\toprule
\multirow{2}{*}{\textbf{How}} & \multicolumn{2}{c}{\textbf{Where}} & \multicolumn{4}{c}{\textbf{Few-shot}} & \multicolumn{2}{c}{\textbf{Full-data}} \\
 \cmidrule(lr){2-3}\cmidrule(lr){4-7}\cmidrule(lr){8-9}
  & Weight by &  Temp. $\gamma$ & 1 & 2 & 5 & $\sim$13 (1\%) & $k$-NN & Linear \\ 
 \midrule
 Base &  & & 40.6 $\pm$ 0.2 & 49.8 $\pm$ 0.2 & 57.9 $\pm$ 0.3 & 63.4 $\pm$ 0.2 & 72.3 $\pm$ 0.1 & 74.4 $\pm$ 0.1\\
 \midrule
\inside & student &  1 & \textbf{41.9 $\pm$ 0.3} & \textbf{51.5 $\pm$ 0.5} & \textbf{59.6 $\pm$ 0.4} & \textbf{65.1 $\pm$ 0.3} & \textbf{73.7 $\pm$ 0.3} & \textbf{75.4 $\pm$ 0.1} \\
\inside-\textsc{Te} & teacher &  1 & 41.5 $\pm$ 0.2 & 50.4 $\pm$ 0.3 & 58.3 $\pm$ 0.3 & 63.7 $\pm$ 0.1 & 72.3 $\pm$ 0.2 & 74.6 $\pm$ 0.1 \\
\inside-\textsc{Max} & student &  0 & - & - & - & - & - & - \\
\inside-\textsc{Max-Te} & teacher &  0 & 41.4 $\pm$ 0.2 & 50.3 $\pm$ 0.3 & 58.1 $\pm$ 0.3 & 63.6 $\pm$ 0.2 & 72.3 $\pm$ 0.2 & 74.5 $\pm$ 0.2 \\
\bottomrule
\end{tabular}
}
\end{center}
\vspace{-\baselineskip}
\end{table}
We present the full table of \cref{table:compare_ens_ablation} that includes all the metrics in \cref{table:compare_ens_ablation_full}. The same observation holds for all metrics. 

For all previous studies, we considered a specific instantiation of \inside strategy, \ie, weight by student predicted probabilities $f_{ijy}=\log s(y|\theta_j,x)$ and $\gamma=1$, which has a nice interpretation of model average (see \cref{loss:ent}). We also studied different variants of the \inside strategy (see \cref{appx:inside_variants}), \begin{itemize}
    \item \inside-\textsc{Te}: weight by teacher $f_{ijy}=\log t_i(y|x)$ and $\gamma=1$; 
    \item \inside-\textsc{Max}: weight by student $f_{ijy}=\log s_j(y|x)$ and $\gamma \to 0$;
    \item \inside-\textsc{Max-Te}: weight by teacher $f_{ijy}=\log t_i(y|x)$ and $\gamma \to 0$
\end{itemize}

\cref{table:compare_ens_prob_variant} compares the downstream performance for all the variants. 
We find that the our \inside (used in our empirical studies) performs better than other variants. 
Interestingly, weighting by the teacher (\inside-\textsc{Te}) performs worse than \inside. We conjecture that this is because the important weights turn out to give a weighted average of teacher predictions as the surrogate target that is \emph{shared} across all students (like \inside) but does not give effective preferential treatment across students which are directly optimized (unlike \inside-\textsc{Te}).
Furthermore, \inside-\textsc{Max} which sharpens the importance weights leads to training divergence. This is probably because the student predictions have higher variance based on which sharp weights lead to unstable training.
In contrast, \inside-\textsc{Max-Te} which uses the (lower-variance) teacher gives reasonable results and comparable to \inside-\textsc{Te}.

\begin{figure}[t]
    \centering
    \begin{subfigure}[b]{.45\textwidth}
        \includegraphics[trim={0 0 0 0}, clip, width=\linewidth]{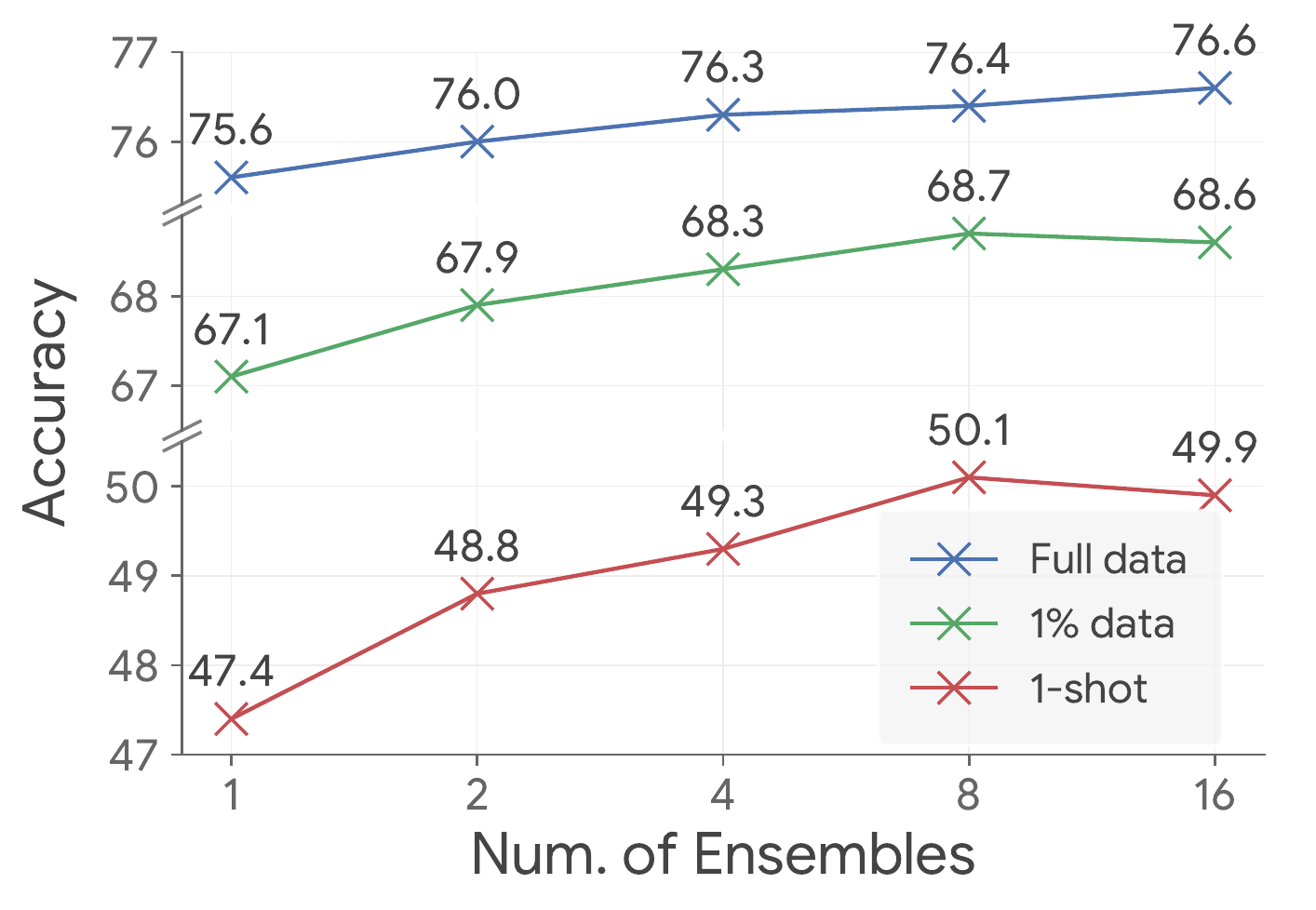}
        \caption{Scaling of ensembles}
        \label{fig:scaling_num_ens_msn}
    \end{subfigure}
    \begin{subfigure}[b]{.45\textwidth}
        \includegraphics[trim={0 0 0 0}, clip, width=\linewidth]{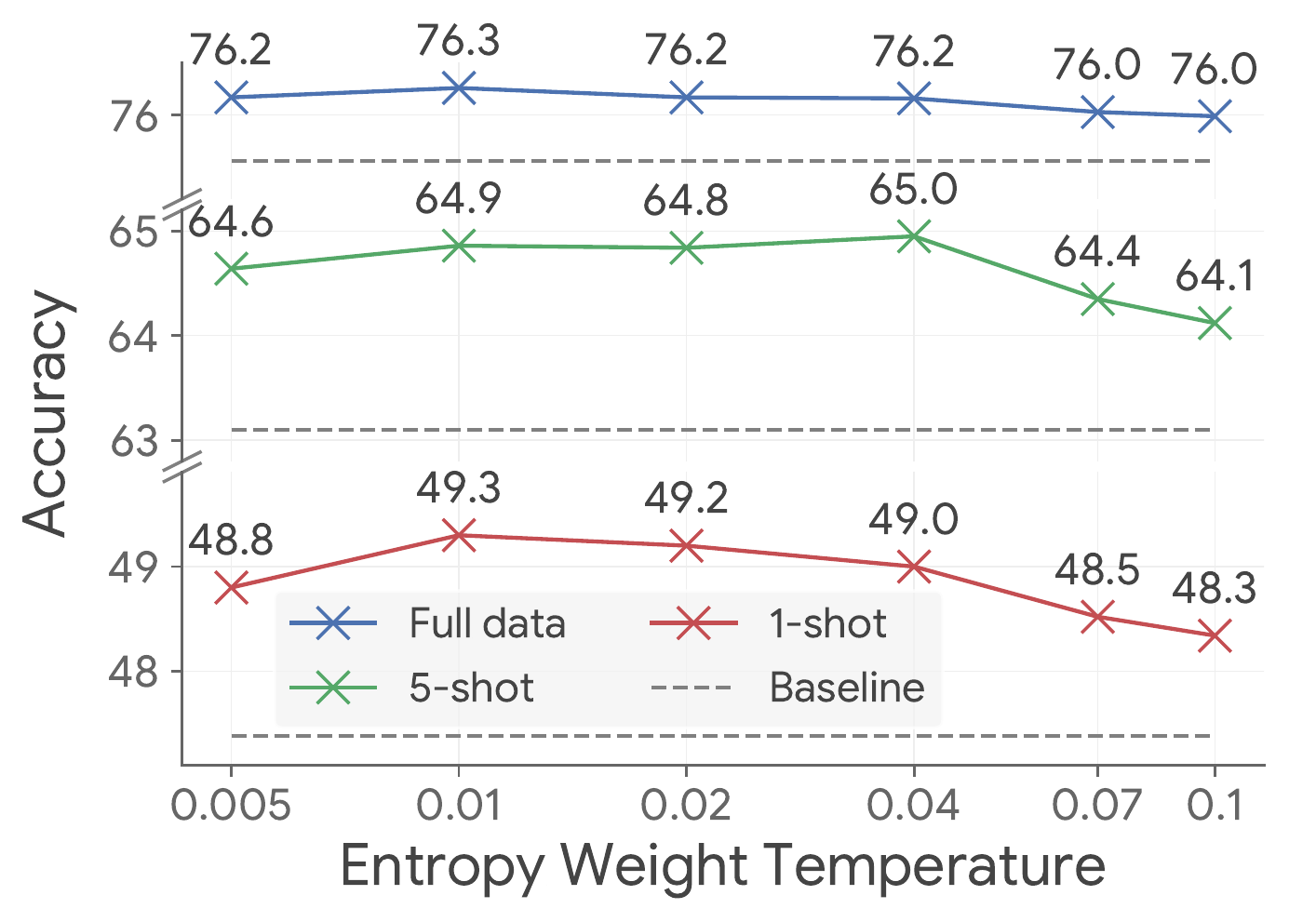}
        \caption{Effect of \ew~temperature $\enttemp$}
        \label{fig:study_ent_temp_msn}
    \end{subfigure}
    \caption{\textbf{Empirical study for \ens{\msnp}{\ew}.} (a) The gains by increasing the number of $(h_\psi,\mu)$-ensembles start to diminish when it is over 8 heads. (b) \msnp prefers smaller temperature for entropy weighting than \dinop.}
\end{figure}

\paragraph{Number of ensembles for \msnp}
In \cref{fig:scaling_num_ens_msn}, we study the effect of increasing the number of $(h_\psi,\mu)$-ensembles for \ens{\msnp}{\ew} with \vits/16 trained for 800 epochs. The scaling trend is similar to \ens{\dinop}{\ew} (\cref{fig:scaling_num_ens_head}) and the gains start to diminish when the number of heads increases above 8.

\paragraph{Effect of \ew temperature $\enttemp$ for \msnp}
\cref{fig:study_ent_temp_msn} studies the effect of entropy weighting temperature $\enttemp$ for \ens{\msnp}{\ew}.
We observed that \msnp is more robust to small temperatures, and the best $\enttemp=0.01$ is smaller than that of \dinop ($\enttemp=0.05$).
When the temperature is too high, the performance drops as a result of under-specialization (\ie, less diversity) as with \dinop.

\subsection{Analyzing $(h_\psi,\mu)$-ensemble diversity}
\paragraph{Visualizing $(h_\psi,\mu)$-ensemble similarity}
\label{appx:vis_head_diversity}
We analyze the diversity between different heads by visualizing the similarity matrix between their codes. Directly measuring the similarity between codes in two heads could not work, because 
1) they may live in different subspaces because of the ensembled projection heads;
2) they may not align in the natural order but in a permuted order.

Therefore, we seek to align codes between different heads by how they are effectively used to \emph{`cluster'} the data. 
In particular, we use a set of randomly sampled inputs $\set{\tinp^i}_{i\in[\numsample]}$ of size $\numsample=51200$ to obtain an empirical code assignment matrix $\assignmat^j \in \R^{\numsample \times \numcode}$ for each $(h_\psi,\mu)$-ensemble $j \in [\numens]$, where the $i$-th row of $\assignmat^j$ corresponds to the teacher predictions 
$t_j(Y|x^i)$.
For the $k$-th code in the head $j$, we extract the $k$-th column from $\assignmat^j$ (\ie, its empirical assignment) as its embedding.
For two codes, we measure their similarity by the cosine similarity between their embeddings.
For a pair of heads $j$ and $j'$, we align their codes using the Hungarian algorithm \citep{kuhn1955hungarian} to maximize the sum of cosine similarity. 
After that, we plot the similarity matrix which is aligned and reordered by the similarity value on the diagonal (in an descending order). 
Note that it is not necessary to do the alignment procedure for the \inside strategy since it is naturally aligned because of the direct distribution averaging over $(h_\psi,\mu)$-ensembles, but we did for fair comparison with other strategies.

We applied the same procedure for different ensemble weighting strategies using \dinop with 4 $(h_\psi,\mu)$-ensembles.
We randomly picked a pair of heads and visualize the similarity matrix before (top row) and after (bottom row) the alignment-reordering setup in \cref{fig:vis_ens_head_diversity}.
We found that before the alignment procedure, the similarity matrix of the \inside strategy already mostly aligns because it explicitly introduces code correspondence between different heads.
Furthermore, by analyzing the similarity decay pattern on the diagonal, it is clear that \ew learns the most diverse $(h_\psi,\mu)$-ensembles while \outside learns the least ones, which may explain the difference of their empirical performance.
For completeness, we also include the visualization of aligned similarity matrices for all pairs of heads in \cref{fig:vis_ens_head_diversity_full_out,fig:vis_ens_head_diversity_full_in,fig:vis_ens_head_diversity_full_ew}, the observations are the same. 

\begin{minipage}{\textwidth}
\begin{figure}[H]
    \centering
    \begin{subfigure}[b]{0.25\linewidth}
        \includegraphics[width=\linewidth, trim={1.5in 0.3in 1.3in 0.4in}, clip]{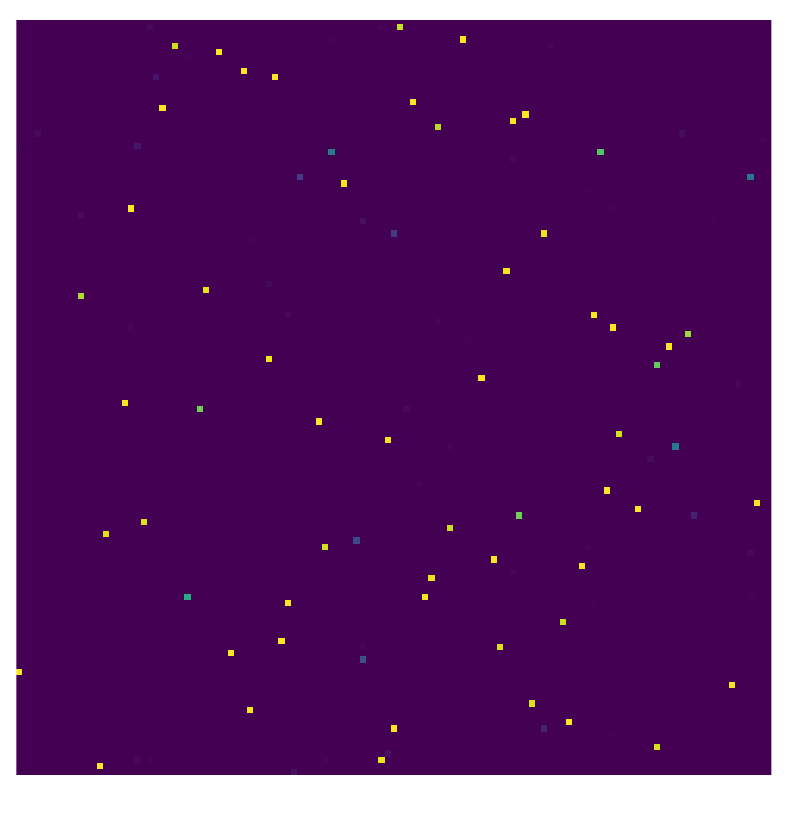}
        \includegraphics[width=\linewidth, trim={1.5in 0.3in 1.3in 0.4in}, clip]{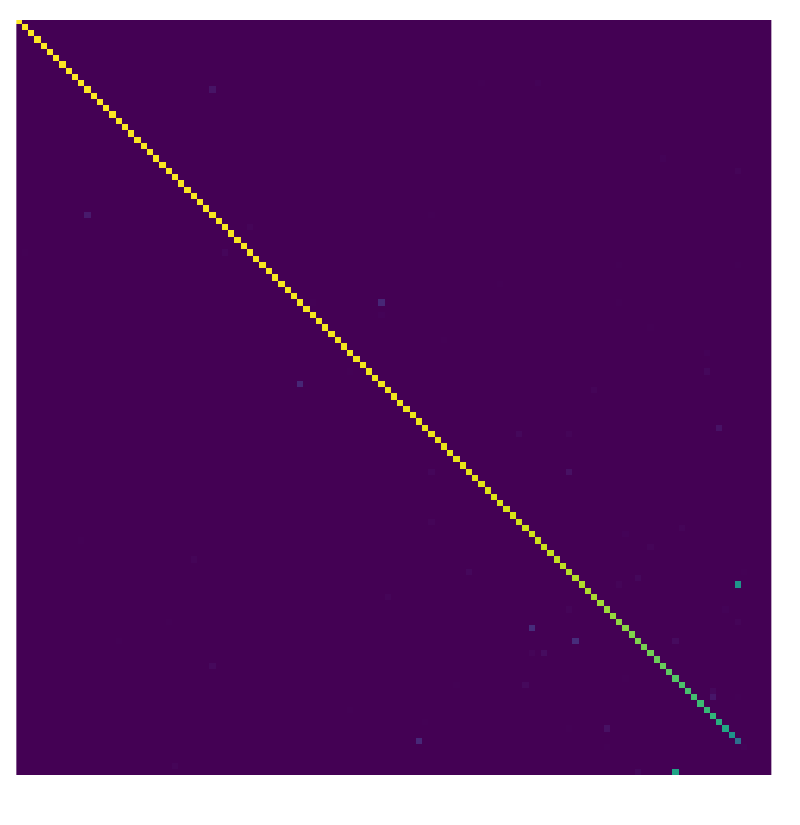}
        \vspace{-\baselineskip}
        \caption{{\outside}}
    \end{subfigure}
    \hspace{1em}
    \begin{subfigure}[b]{0.25\linewidth}
        \includegraphics[width=\linewidth, trim={1.5in 0.3in 1.3in 0.4in}, clip]{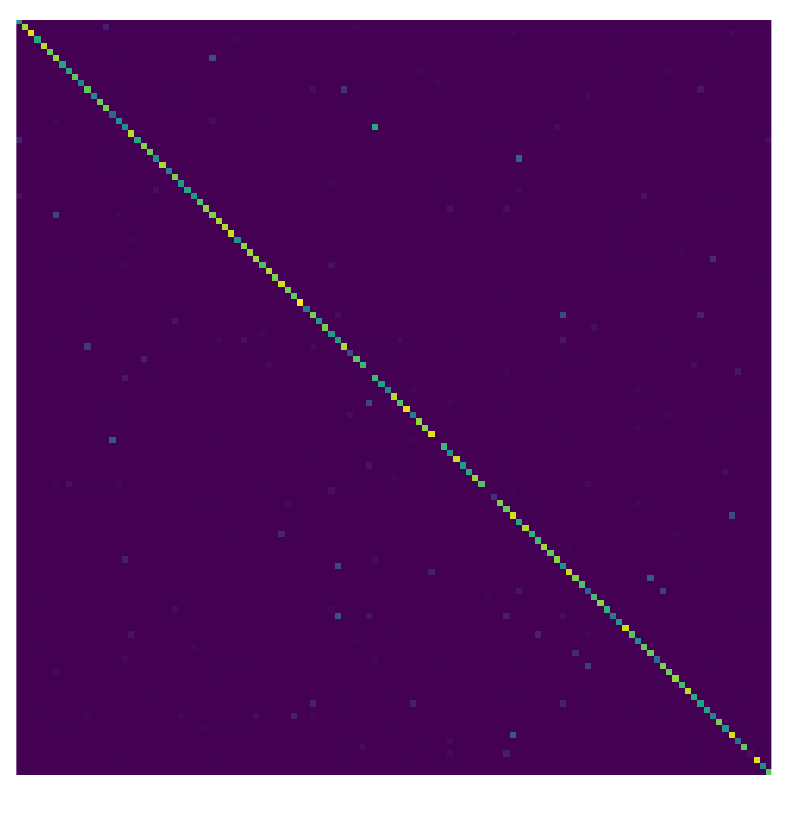}
        \includegraphics[width=\linewidth, trim={1.5in 0.3in 1.3in 0.4in}, clip]{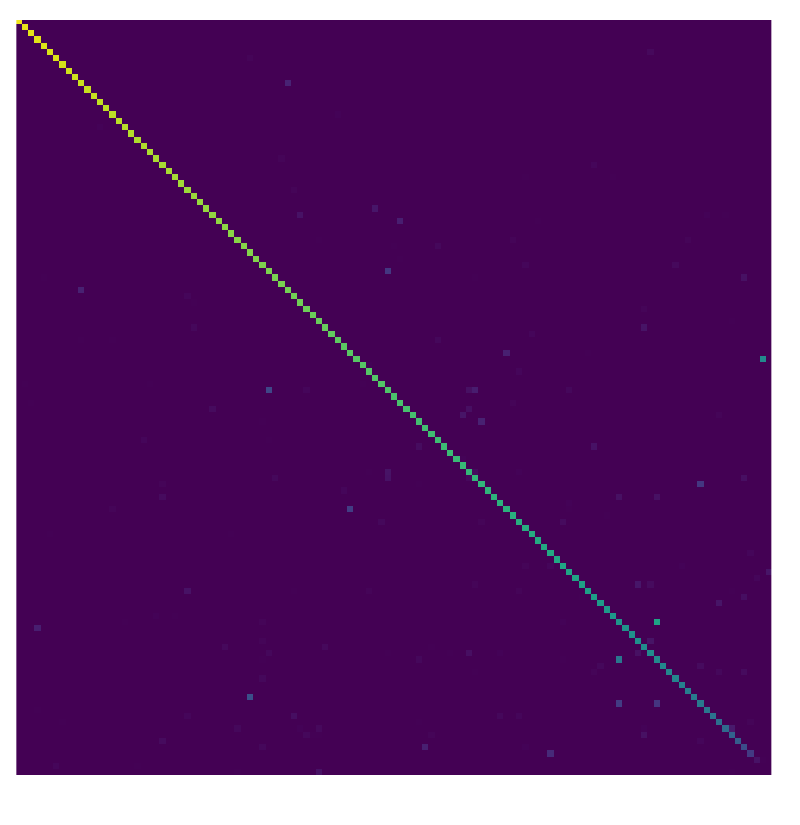}
        \vspace{-\baselineskip}
        \caption{{\inside}}
    \end{subfigure}
    \hspace{1em}
    \begin{subfigure}[b]{0.295\linewidth}
        \includegraphics[width=0.85\linewidth, trim={1.5in 0.3in 1.3in 0.4in}, clip]{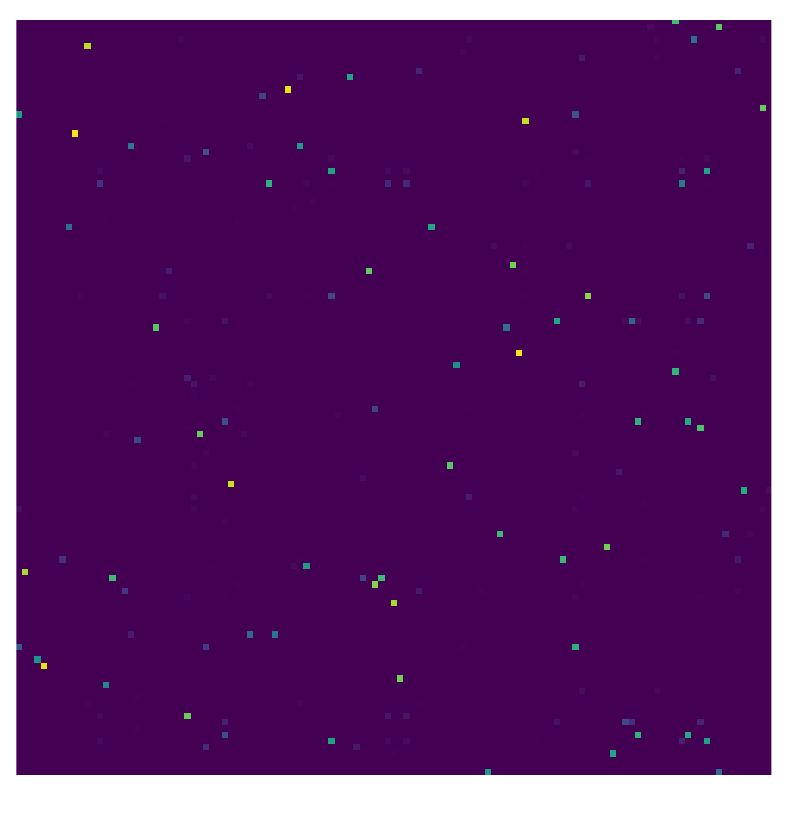}
        \includegraphics[width=\linewidth, trim={1.4in 0.3in 0.8in 0.4in}, clip]{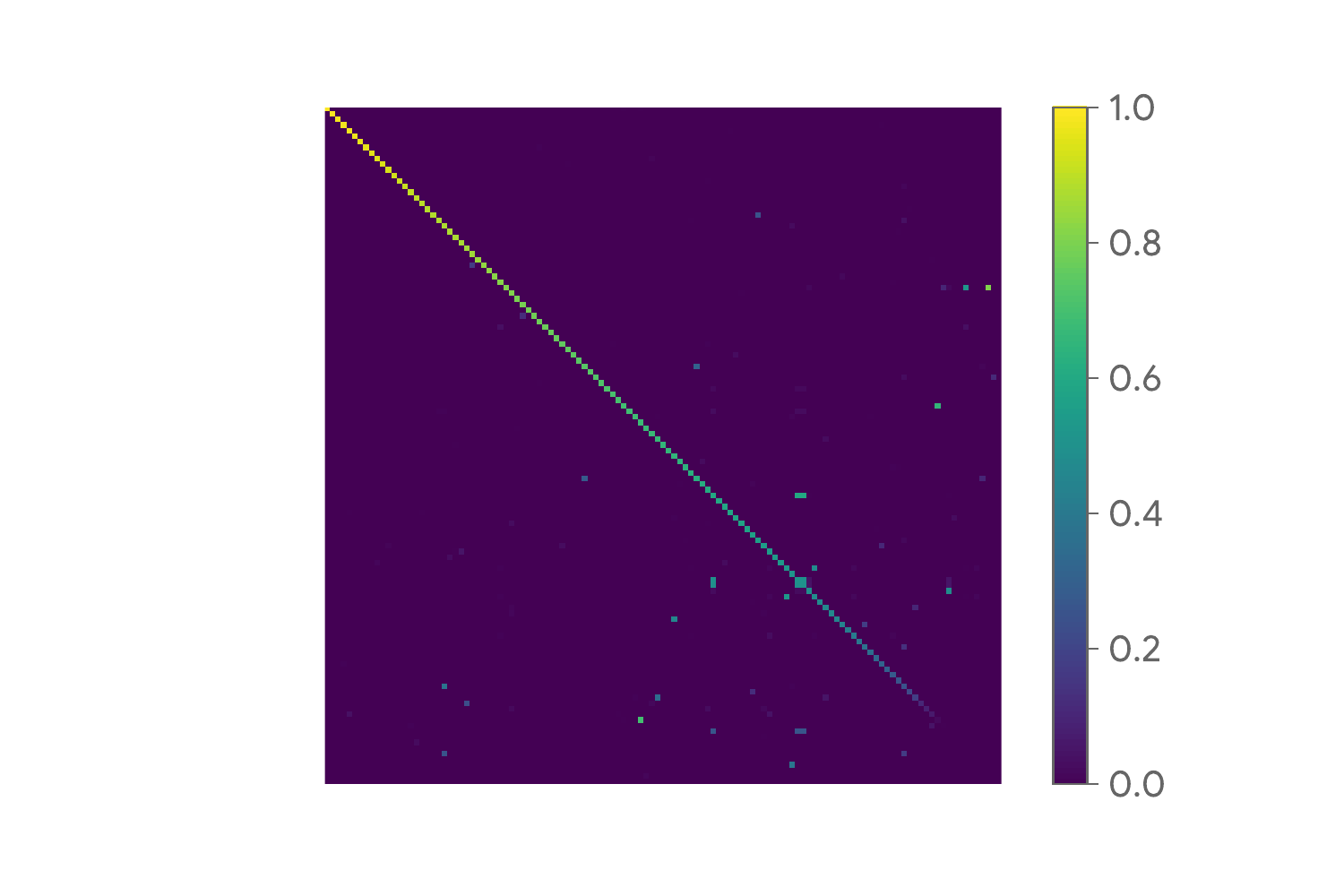}
        \vspace{-\baselineskip}
        \caption{{\ew}}
    \end{subfigure}
    \caption{\textbf{Visualization of $(h_\psi,\mu)$-ensemble diversity.} 
    \ew learns the most diverse $(h_\psi,\mu)$-ensembles while \outside learns the least ones.
    We visualize the code similarity matrix between a pair of randomly selected projection heads.
    Top row shows the original similarity matrix (\ie, in natural order) and the bottom row shows the aligned similarity matrix which aligns codes by empirical assignment probabilities.
    \dinop \vits/16 with 4 heads is used.
    Best viewed in color.}
    \label{fig:vis_ens_head_diversity}
\end{figure}%
\vspace{-2\baselineskip}
\begin{figure}[H]
    \centering
    \includegraphics[trim={1in 2.5in 0.5in 1.5in}, clip, width=0.9\linewidth]{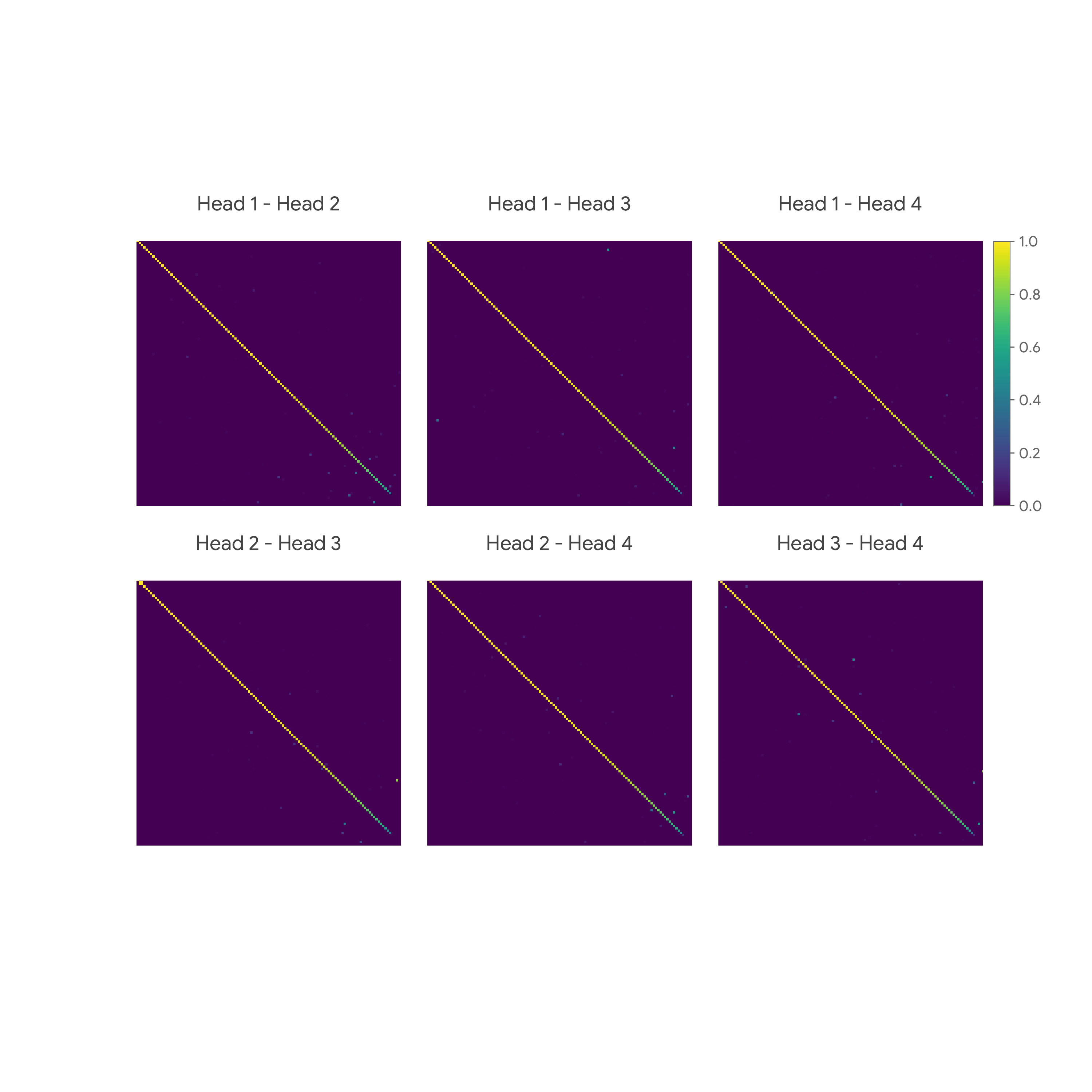}
    \caption{\textbf{Visualization of $(h_\psi,\mu)$-ensemble diversity between all pairs of heads for \ens{\dinop}{\outside}.}
    The \outside strategy does not learn diverse $(h_\psi,\mu)$-ensembles.
    \dinop with \vits/16 and 4 heads is used. Best viewed in color.}
    \label{fig:vis_ens_head_diversity_full_out}
\end{figure}
\end{minipage}
\newpage
\begin{minipage}{\textwidth}
\begin{figure}[H]
    \centering
    \includegraphics[trim={1in 2.5in 0.5in 1.5in}, clip, width=0.9\linewidth]{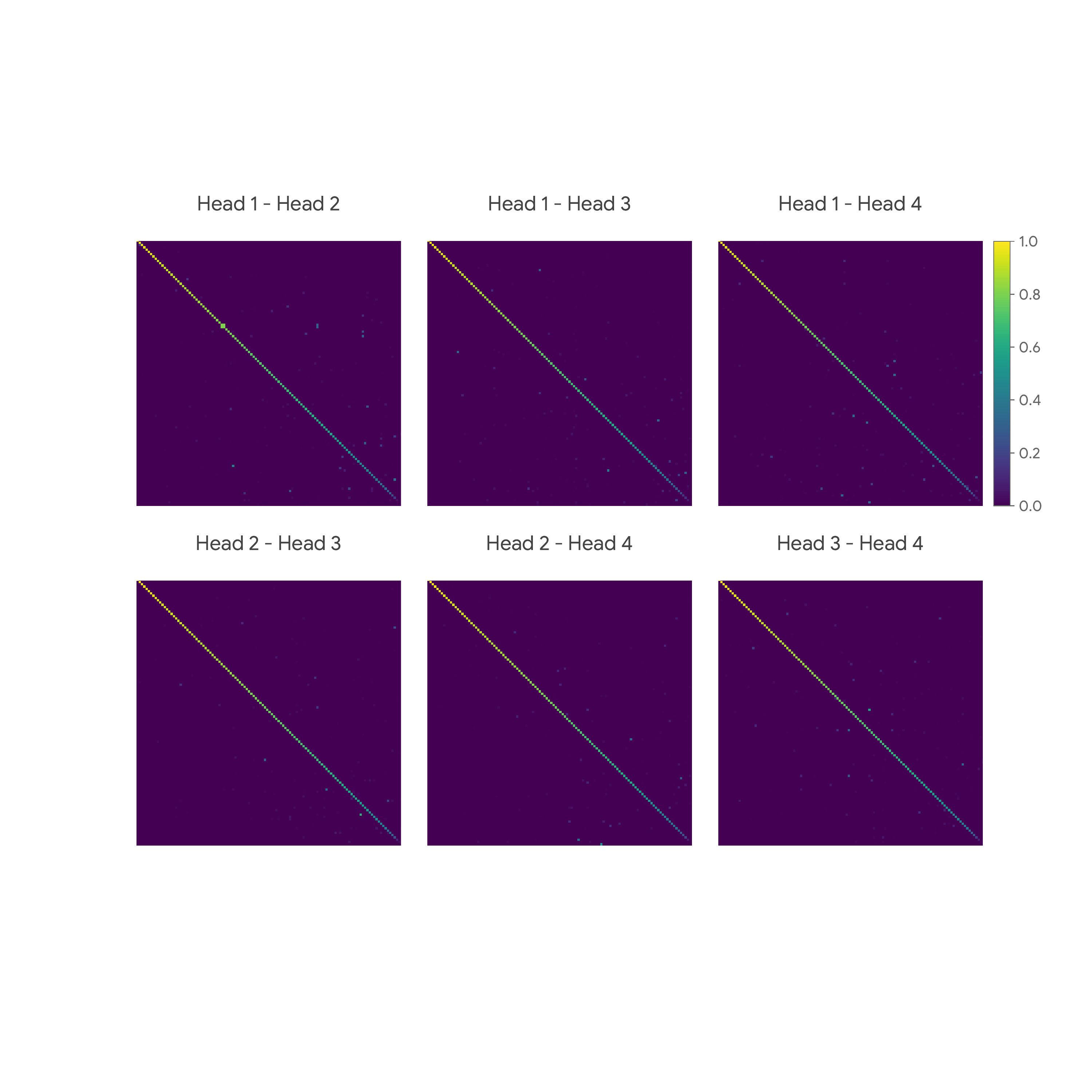}
    \caption{\textbf{Visualization of $(h_\psi,\mu)$-ensemble diversity between all pairs of heads for \ens{\dinop}{\inside}.} The \inside strategy learns more diverse $(h_\psi,\mu)$-ensembles than \outside.
    \dinop with \vits/16 and 4 heads is used. Best viewed in color.}
    \label{fig:vis_ens_head_diversity_full_in}
\end{figure}%
\vspace{-2\baselineskip}
\begin{figure}[H]
    \centering
    \includegraphics[trim={1in 2.5in 0.5in 1.5in}, clip, width=0.9\linewidth]{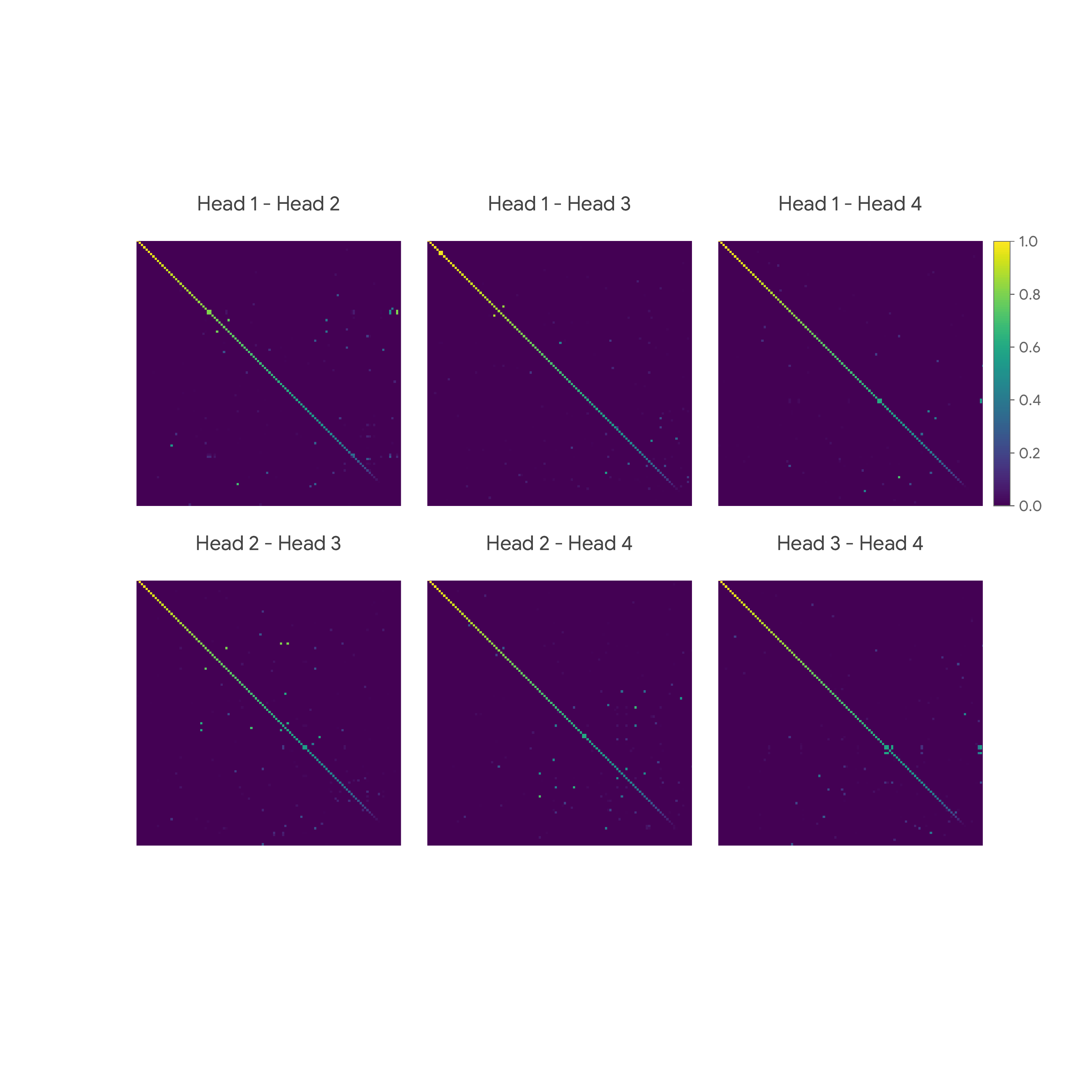}
    \caption{\textbf{Visualization of $(h_\psi,\mu)$-ensemble diversity between all pairs of heads for \ens{\dinop}{\ew}.} The \ew strategy learns the most diverse $(h_\psi,\mu)$-ensembles.
    \dinop with \vits/16 and 4 heads is used. Best viewed in color.}
    \label{fig:vis_ens_head_diversity_full_ew}
\end{figure}
\end{minipage}

\section{Analysis}\label{app:analysis}

\subsection{Derivations}
\label{appx:derivations}
In this subsection, we provide derivations for some non-trivial losses that we explore within our framework.

Recall that our weighted cross-entropy loss is of the form,
\begin{align}
\mathcal{L}_n(\theta)&= \frac{1}{n}\sum_{x\in\mathcal{D}_n} \sum_{i,j\in[m]}  \HH^{\times}[w_{ijY} \odot t_i(Y|x),s(Y|\theta_j,x)]\\
&= \frac{1}{n}\sum_{x\in\mathcal{D}_n} \sum_{i,j\in[m]}  \sum_{y \in \Y} w_{ijy} t_i(y|x) \log s(y|\theta_j,x)\\
\text{where}\quad w_{ijy}&=\softmax\left(\left\{\tfrac{1}{\gamma}f_{ijy}(\stopgrad(\theta),x) :i,j\in[m]\right\}\right).
\end{align}

Fuethermore, observe that,
\begin{align}
\nabla_{\theta} \sum_{i,j\in[m]} \HH^\times[w_{ijY}\odot t_i(Y|x), s(Y|\theta_j,x)]
&= \sum_{i,j\in[m]} \int_\Y w_{ijy} t_i(y|x)\nabla_{\theta}\log s(y|\theta_j,x)  \dd y.
\end{align}
This indicates that the proposed weighted ensemble SSL loss is simply a reweighted log-likelihood loss. We use this fact in our derivation of probability weighting (\inside) loss.

\paragraph{Uniform weighting (\outside)} 
Our \outside strategy in \cref{eq:outside} 
uses $f_{ijy}=\log \delta(i-j)$ which gives $w_{ijy}=\frac{1}{m}\delta(i-j)$ (for any choice of $\gamma$), thus the loss,
\begin{align}
     \mathcal{L}_n^\text{\outside}(\theta)
      &= \frac{1}{n}\sum_{x\in\mathcal{D}_n} \sum_{i,j\in[m]}  \sum_{y \in \Y} \frac{1}{m}\delta(i-j) t_i(y|x) \log s(y|\theta_j,x) \\
      &=\frac{1}{n} \sum_{x\in\mathcal{D}_n} \frac{1}{m}\sum_{i\in[m]} 
    \HH^\times[t_i(Y|x), s(Y|\theta_i,x)]
\end{align}
This loss assigns equal weights to $m$ pairs of pairwised student/teacher.

An straightforward generalization is to assign equal weights to all possible pairs ($m^2$) of student/teacher with $f_{ijy}=0$ and $w_{ijy}=\frac{1}{m^2}$, which gives the \outside-\textsc{All} loss,
\begin{align}
\mathcal{L}^\text{\outside-\textsc{All}}_n(\theta)=\frac{1}{n}\sum_{x\in\mathcal{D}_n} \frac{1}{m^2}\sum_{i,j\in[m]} \HH^\times\left[ t_i(Y|x), s(Y|\theta_j,x)\right],
\end{align}

\paragraph{Probability weighting (\inside)} 
\label{appx:inside_variants}
Recall our \inside loss in \cref{eq:inside} has the form,
\begin{align}
  \mathcal{L}^\text{\inside}_n(\theta) = \frac{1}{n} \sum_{x\in\mathcal{D}_n}  
    \HH^\times\left[ \frac{1}{m} \sum_{i\in[m]} t_i(Y|x), \frac{1}{m}\sum_{j\in[m]} s(Y|\theta_j,x)\right].
\end{align}
We derive its equivalence with our general loss with $f_{ijy}=\log s(y|\theta_j,x)$ and $\gamma=1$ in terms of the gradients,
\begin{align}
\nabla_\theta\mathcal{L}^\text{\inside}_n(\theta)
&=\frac{1}{m}\sum_{i\in[m]} \int_\Y t_i(y|x) \log \frac{1}{m} \sum_{j\in[m]} s(y|\theta_j,x) \dd y\\
&=\frac{1}{m}\sum_{i\in[m]} \int_\Y t_i(y|x) \nabla_\theta \log \frac{1}{m} \sum_{j\in[m]} s(y|\theta_j,x) \dd y\\
&=\frac{1}{m}\sum_{i\in[m]} \int_\Y t_i(y|x) \frac{\frac{1}{m} \sum_{j\in[m]} \nabla_\theta s(y|\theta_j,x)}{\frac{1}{m} \sum_{j\in[m]} s(y|\theta_j,x)} \dd y\\
&=\frac{1}{m}\sum_{i\in[m]} \int_\Y t_i(y|x) \frac{\frac{1}{m} \sum_{j\in[m]} s(y|\theta_j,x)\nabla_\theta \log s(y|\theta_j,x)}{\frac{1}{m} \sum_{j'\in[m]} s(y|\theta_{j'},x)} \dd y\\
&=\frac{1}{m}\sum_{i,j\in[m]} \int_\Y t_i(y|x) \frac{ s(y|\theta_j,x)}{\sum_{j'\in[m]} s(y|\theta_{j'},x)} \nabla_\theta \log s(y|\theta_j,x)\dd y\\
&=\nabla_\theta\frac{1}{m}\sum_{i,j\in[m]}  \HH^\times\left[w_{ijY}\odot t_i(Y|x), s(Y|\theta_j,x) \right]
\end{align}
where $w_{ijy} = \frac{ s(y|\theta_j,x)}{\sum_{j'\in[m]} s(y|\theta_{j'},x)}$ (or equivalently, $f_{ijy}=\log s(y|\theta_j,x)$ and $\gamma=1$). The last equality is because $w_{ijy}$ is stopped gradient with respect to $\theta$. 
This is the same analysis as done in \citet{burda2016iwae}.
The above formation establishes the equivalence of gradients between two losses, which implies the same behavior (\eg, optimum) using gradient-based optimization, as the common practice of deep learning.

We also generalize this loss to some variants which we explore in \cref{table:compare_ens_prob_variant}.
A ``dual'' variant is to use teacher predictions $f_{ijy}=\log t_i(y|x)$ instead of student ones; this implies $w_{ijy} = \frac{ t_i(y|x)}{\sum_{i'\in[m]} t_{i'}(y|x)}$ and the \inside-\textsc{Te} loss,
\begin{align}
  \mathcal{L}^\text{\inside-\textsc{Te}}_n(\theta) = \frac{1}{n}\sum_{x\in\mathcal{D}_n} \sum_{i,j\in[m]}  \sum_{y \in \Y} \frac{ t_i(y|x)}{\sum_{i'\in[m]} t_{i'}(y|x)} t_i(y|x) \log s(y|\theta_j,x).
\end{align}
Note that this simply reduces to use a weighted teacher predictions $\frac{ t_i(y|x)}{\sum_{i'\in[m]} t_{i'}(y|x)} t_i(y|x) $ as the surrogate target that is shared across all students.

Another generalization is to use ``hard'' weighting, \ie, $\gamma \to 0$, which gives the \inside-\textsc{Max} loss that only assigns weight to the most confident student,
\begin{align}
\mathcal{L}_n^\text{\inside-\textsc{Max}}(\theta)
&= \frac{1}{n}\sum_{x\in\mathcal{D}_n} \sum_{i,j\in[m]}  \sum_{y \in \Y} w_{ijy} t_i(y|x) \log s(y|\theta_j,x)\\
\text{where}\quad w_{ijy} &=  \delta(i-i^*)\delta(j-j^*),\quad (i^*,j^*)=\arg\max_{ij}f_{ijy}, \forall y \in \Y.
\end{align}
This loss reduces to a generalization of multiple choice learning \citep{guzman2012mcl} used in multi-headed networks \citep{lee2015multihead} in our ensemble SSL setup.
Similarly we can also derive the dual variant of it that uses the teacher predictions, which is omitted here for brevity.

\paragraph{Entropy weighting (\ew)} The derivation of \ew loss in \cref{eq:ent_weight} is similar to the \outside loss but applies an entropy weights. Recall that we use $f_{ijy}=-\HH[t_i(Y|x)]+\log \delta(i-j)$, which gives $w_{ijy}=\softmax_i(\{-\frac{1}{\gamma}\HH[t_{i'}(Y|x)]:i'\in[m]\})$ and,
\begin{align}
    \mathcal{L}^\text{\ew}_n(\theta) &= \frac{1}{n} \sum_{x\in\mathcal{D}_n}  
    \sum_{i\in[m]} \softmax_i(\{-\tfrac{1}{\gamma}\HH[t_{i'}(Y|x)]:i'\in[m]\}) \HH^\times\left[ t_i(Y|x), s(Y|\theta_i,x)\right]. 
\end{align} 
One can also generalizes it to its dual variant which uses the student entopies, \ie,$f_{ijy}=-\HH[s(Y|\theta_j, x)]+\log \delta(i-j)$, which gives the \ew-\textsc{St} loss,
\begin{align}
    \mathcal{L}^\text{\ew-\textsc{St}}_n(\theta) &= \frac{1}{n} \sum_{x\in\mathcal{D}_n}  
    \sum_{i\in[m]} \softmax_i(\{-\tfrac{1}{\gamma}\HH[s(Y| \theta_{i'}, x)]:i'\in[m]\}) \HH^\times\left[ t_i(Y|x), s(Y|\theta_i,x)\right]. 
\end{align}

\subsection{Relating some losses}
Here, we relate some losses derived above. Specifically, we relate the uniform weighting (\outside, \outside-\textsc{All}) and probability weighting (\inside) in \cref{appx:loss_relate_in_out}, and relate entropy weighting (\ew) and variance weighting in \cref{appx:loss_relate_ent_var}.

\subsubsection{Uniform \& Probability Weighting}
\label{appx:loss_relate_in_out}
We first establish the relation between \outside and \inside using the joint convexity of unnormalized KL divergence and the fact that our weighted cross-entropy loss is a weighted unnormalized KL divergence up to some constant in $\theta$.
In particular, the joint convexity of unnormalized KL divergence can be shown by combining the facts that Csisz\`ar $f$-divergences are jointly convex (Proposition 1 in \citet{dragomir2013generalization}) and unnormalized KL divergence corresponds to the convex generator, $f(u)=u\log u -u + 1$, as required by the proposition.

First, our weighted cross-entropy loss is unnormalized KL divergence up to some constant in $\theta$:
\begin{align}
\mathcal{L}_n^\text{\outside}(\theta)
   &=\frac{1}{n} \sum_{x\in\mathcal{D}_n} \frac{1}{m}\sum_{i\in[m]} 
 \KK[t_i(Y|x), s(Y|\theta_i,x)] + \text{constant}\\
\mathcal{L}^\text{\inside}_n(\theta)
&=\frac{1}{n}\sum_{x\in\mathcal{D}_n}  \KK\left[\frac{1}{m}\sum_{i\in[m]} t_i(Y|x), \frac{1}{m}\sum_{j\in[m]} s(Y|\theta_j,x)\right] + \text{constant}
\end{align}
Therefore, the joint convexity of (unnormalized) KL divergence directly implies an ordering of the loss up to some constant in $\theta$, i.e.,
\begin{align}
&\mathcal{L}^\text{\inside}_n
\le \mathcal{L}^\text{\outside}_n
\end{align}
Furthermore, we can also relate \inside and \outside-\textsc{All} using the fact that the (unnormalized) cross-entropy $\HH^\times[p(X),q(X)]$ is linear in the first argument $p$ but convex in the second argument $q$, which implies,
\begin{align}
    &\mathcal{L}^\text{\inside}_n \le \mathcal{L}^\text{\outside-\textsc{All}}_n
\end{align}

\subsubsection{Entropy \& Variance Weighting}
\label{appx:loss_relate_ent_var}

Suppose $p(X)$ is a discrete distribution (normalized) on $\X=[c]$. It can be shown that,
\begin{align}
\HH[p(X)]\le
\tfrac{1}{2} \log\left(\Var_p[X]+\tfrac{1}{12}\right)
+
\tfrac{1}{2} \log\left(2\pi e\right)
\end{align}
where $\Var_p[X]=\sum_{x\in[c]} p(x)(x-\mu)^2$ and $\mu=\EE_p[X]=\sum_{x\in[c]}p(x)x$ (Theorem 9.7.1, \citet{cover1999elements}). Note, a tighter bound \cite{mow1998tight} also exists but it places stronger restrictions on $p$.
This relationship suggests that choosing weights proportional to $\exp(-\HH[t_i(Y|x)])$ (as in \ew) is potentially related to choosing weights proportional to weighting by variance 
$(\Var_{t_i(Y|x)}[Y]+\epsilon)^{-1/2}$ where $(\epsilon=\frac{1}{12})$.

\end{document}